\documentclass[10pt,twocolumn,letterpaper]{article}

\usepackage{cvpr}              

\usepackage{graphicx}%
\usepackage{multirow}%
\usepackage[section]{placeins}
\usepackage{makecell}

\usepackage[dvipsnames]{xcolor}

\definecolor{cvprblue}{rgb}{0.21,0.49,0.74}
\usepackage[pagebackref,breaklinks,colorlinks,citecolor=cvprblue]{hyperref}

\def\paperID{2574} 
\def\confName{CVPR}
\def\confYear{2025}

\def\ModelName{VideoChat-TPO}
\def\MethodName{Task Preference Optimization}

\def\unfreeze{{\includegraphics[height=0.8\baselineskip, width=0.6\baselineskip]{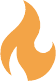}}}
\def\freeze{{\includegraphics[height=0.7\baselineskip, width=0.7\baselineskip]{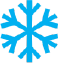}}}

\usepackage{tcolorbox}

\title{Task Preference Optimization: Improving Multimodal Large Language Models with Vision Task Alignment}

\author{Ziang Yan$^{*2,1}$, Zhilin Li$^{*3,1}$, Yinan He$^{*1}$ \\
Chenting Wang$^{4,1}$, Kunchang Li$^{5,1}$, Xinhao Li$^{6,1}$, Xiangyu Zeng$^{6,1}$\\
Zilei Wang$^{3}$, Yali Wang$^{5,1}$,Yu Qiao$^{1}$, Limin Wang$^{6,1}$, Yi Wang$^{\dagger 1,7}$\\
\small{
$^1$Shanghai AI Laboratory\quad 
$^2$Zhejiang University \quad 
$^3$University of Science and Technology of China \quad $^4$Shanghai Jiao Tong University}\\
\small{
$^5$Shenzhen Institutes of Advanced Technology, Chinese Academy of Sciences\quad 
$^6$Nanjing University \quad 
$^7$Shanghai Innovation Institute
} \\ 
\small{\url{https://github.com/OpenGVLab/TPO}} \\
}
\begin{document}
\maketitle
\begin{abstract}
Current multimodal large language models (MLLMs) struggle with fine-grained or precise understanding of visuals although they give comprehensive perception and reasoning in a spectrum of vision applications. Recent studies either develop tool-using or unify specific visual tasks into the autoregressive framework, often at the expense of overall multimodal performance. To address this issue and enhance MLLMs with visual tasks in a scalable fashion, we propose Task Preference Optimization (TPO), a novel method that utilizes differentiable task preferences derived from typical fine-grained visual tasks.
TPO introduces learnable task tokens that establish connections between multiple task-specific heads and the MLLM. By leveraging rich visual labels during training, TPO significantly enhances the MLLM's multimodal capabilities and task-specific performance.
Through multi-task co-training within TPO, we observe synergistic benefits that elevate individual task performance beyond what is achievable through single-task training methodologies.
Our instantiation of this approach with VideoChat and LLaVA demonstrates an overall 14.6\% improvement in multimodal performance compared to baseline models. Additionally, MLLM-TPO demonstrates robust zero-shot capabilities across various tasks, performing comparably to state-of-the-art supervised models. 

{
  \renewcommand{\thefootnote}%
    {\fnsymbol{footnote}}
  \footnotetext[0]{*Equal contribution. $\dagger$Corresponding author.} 
  }
\end{abstract}    
\section{Introduction}
\label{sec:intro}

 \begin{figure}[ht]
    \centering
    \includegraphics[width=\linewidth]{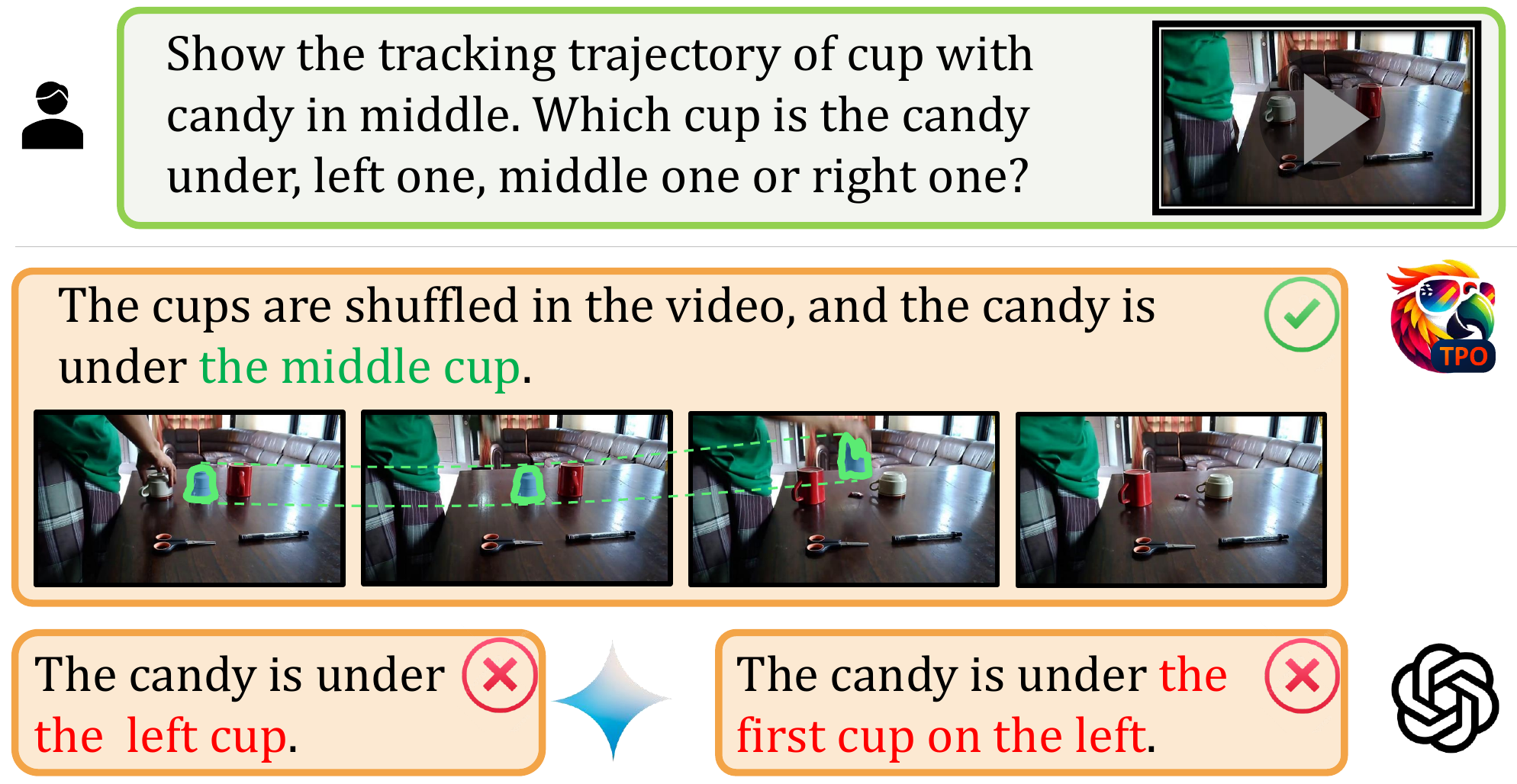}
    \caption{TPO uses differentiable task preferences from dense visual supervisions via task-specific heads to enhance MLLMs in fine-grained understanding.
    }
    \label{fig:visualize}
    \vspace{-3.4mm}
\end{figure}

Multimodal large language models (MLLMs) show impressive visual perception and reasoning capabilities, with applications in personal assistants~\cite{team2024gemini}, embodied systems~\cite{driess2023palm}, and scientific discovery~\cite{chen2023fengwu}. Considering the growing expectations of users in more accurate and detailed perception, taking the shell game as an example, further improving the generality of MLLM requires fine-grained knowledge representation beyond words, \eg MLLM implicitly embeds how to track keys indicated by users.

Existing studies address the enhancement of perceptual granularity in MLLMs by focusing on specific visual tasks (\eg temporal grounding, segmentation, tracking) via MLLMs. They usually fine-tune MLLMs on more task data in text format or enable MLLMs to activate the corresponding task heads. 
Shikra~\cite{chen2023shikra} applies MLLM to localization tasks, transforms object coordinates into dialogue formats, and learns them autoregressively. Meanwhile, TimeChat~\cite{Ren2023TimeChat} and VideoChat-T~\cite{zengtimesuite} treat event timestamps as text for autoregressive prediction, thereby endowing MLLMs with temporal grounding.
They do improve specific task performance significantly, while at the cost of multimodal performance more or less. This is counterintuitive as seminal research proves different visual tasks are correlated and training them together often yields mutual gains~\cite{intern,univtg,jiang2023sparse}. 
We conjecture that the presentation of different tasks influences this training and that the conflict arises from the learning discrepancies between discrete textual tokens and visually dense predictions.
Our experiments in Section~\ref{experiment} validate that a decoupled representation design can effectively address this issue.

\begin{figure}[t]
    \centering
    \includegraphics[width=\linewidth]{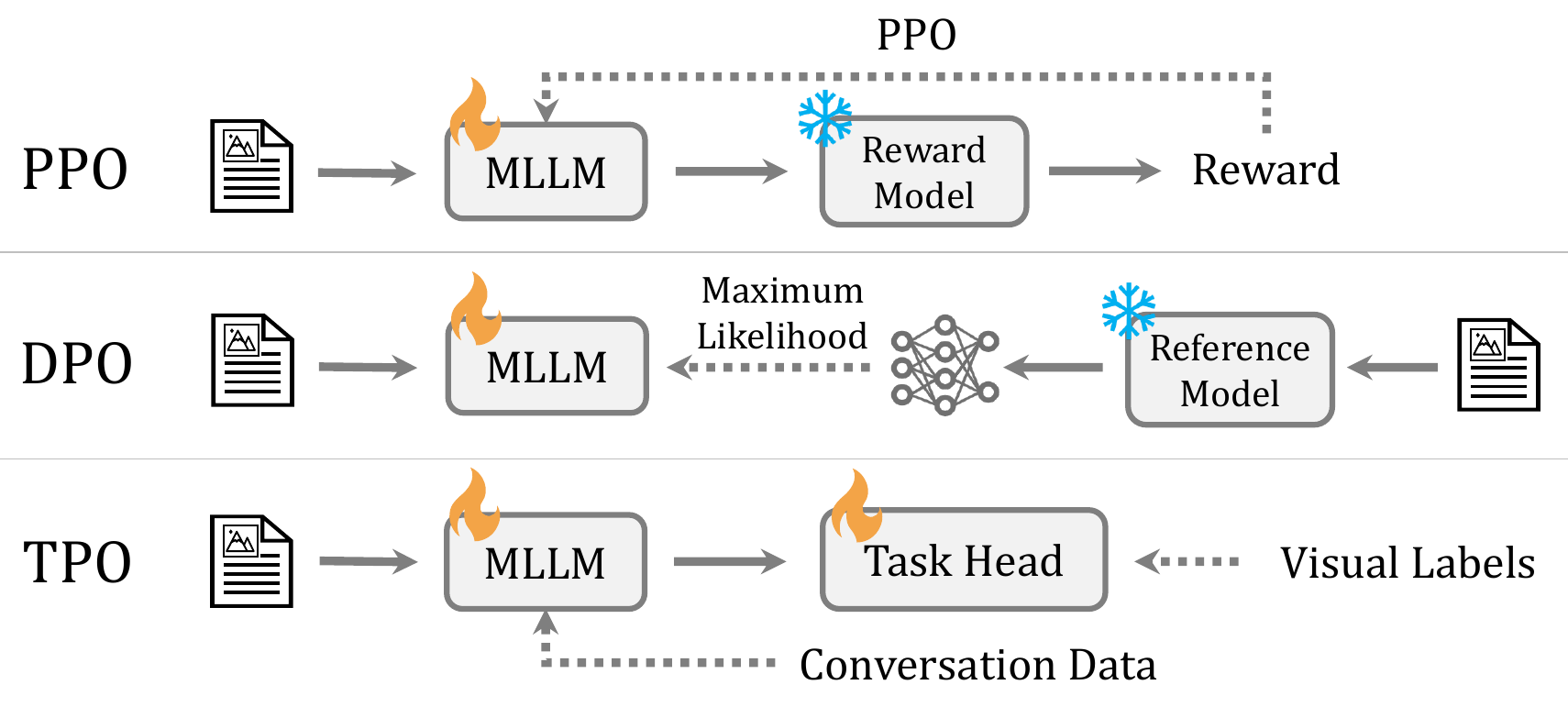}
    \caption{\textbf{Comparison of Learning Method.} A solid line indicates data flow, and a dotted line represents feedback.
    \freeze{} and \unfreeze{} denote modules that are frozen and unfrozen.
    }
    \label{fig:intro}
    \vspace{-4mm}
\end{figure}

To enhance the multimodal capabilities of MLLMs, we explore optimization methods to meet multiple visual task requirements in an end-to-end manner.
We propose task preference optimization (TPO), incorporating visual task knowledge into the MLLMs by jointly maximizing the likelihoods of visual task estimations and multimodal dialogue. 
Inspired by direct preference optimization~\cite{rafailov2024direct} (DPO) and related methods, we treat visual task annotations as human preferences in particular demands, as shown in Figure ~\ref{fig:intro}. DPO aids LLM (or MLLM) in aligning with human preferences through a binary classification that directs the model to generate responses that people prefer. Similarly, TPO enhances MLLMs' visual sensing capabilities through differentiable task optimizations that guide MLLMs to yield dense predictions closely resembling human perception. To achieve this, TPO concretizes MLLM-specific visual perceptions into corresponding task tokens, disentangled from MLLM representation. Then TPO fine-tunes these task tokens and updates the MLLM accordingly.

Specifically, TPO appends visual task heads to the partial output of the MLLM, using several learnable task tokens as inputs for these corresponding heads. During training, TPO enables the MLLM to first distinguish and activate the appropriate task tokens based on user instructions. 
Subsequently, TPO jointly trains the task tokens and their corresponding heads to enhance the MLLM's understanding of visual tasks. Finally, TPO trains the entire model—including task tokens and heads—on both multimodal and visual-specific task data, promoting the perception and reasoning capabilities of the MLLM. Additionally, we note that multi-task co-training yields greater improvements than single-task training.

TPO demonstrates scalability across various MLLM approaches, encompassing a wide range of visual task categories and data quantities. 

We validate the effectiveness of TPO within widely used MLLMs, such as LLaVA~\cite{llava} and VideoChat~\cite{videochat}, as detailed in Section~\ref{ablation}. By fine-tuning these open-source MLLMs with TPO, we enhance their visual understanding capabilities and improve dialogue performance.
Additionally, we explore several key spatiotemporal and spatial perception tasks, including spatial grounding, tracking, and temporal grounding.
Our findings indicate that these tasks can mutually enhance each other's performance, particularly contributing to the improvement of multimodal dialogue capabilities. 

Our contributions can be summarized as:
\begin{itemize} 
    \item We propose a new training method for multimodal large language models, referred to as  \MethodName{} (TPO). This method leverages supervised information from visual task-specific data to optimize the MLLM through the corresponding heads, resulting in significantly enhanced multimodal perception and reasoning performance. Specifically, TPO achieves an average improvement of 14.6\%  across multiple image and video multimodal benchmarks~\cite{videomme,mvbench,MLVU,meng2024mmiu, seedbench2}.
    \item TPO effectively equips MLLM with the capability to address several key visual perception tasks through the introduced task heads. MLLM-TPO achieves comparable performance in spatial grounding, moment retrieval, highlight detection, referring segmentation, and tracking comparable to expert models across various benchmarks.
    \item TPO demonstrates scalability across various employed MLLMs, task heads, and scales of task data. We validate the effectiveness of TPO in multiple mainstream MLLMs, such as VideoChat2~\cite{mvbench} and LLaVA~\cite{llava, Llava-onevision}. Notably, multi-task joint training based on TPO enhances both multimodal performance and individual visual task, with improvements becoming increasingly significant as additional appropriate heads are introduced. Furthermore, the performance of the MLLM and task heads improves when scaling task data.
\end{itemize}
\vspace{-0.5mm}

\section{Related Work}

\paragraph{Vision Foundation Model.}
Vision foundation models~\cite{videoclip,intern,li2023unmasked, videoprism, wang2023masked, videococa, videomae, wang2023videomae, wang2024internvideo2, tsn} are designed to be adaptable for various downstream tasks through extensive pre-training on large-scale and diverse datasets. 
VideoPrism~\cite{videoprism} achieves the leading results in various video tasks by combining video-text contrastive learning and spatiotemporal token reconstruction using a dataset of public and proprietary videos.
InternVideo2~\cite{wang2024internvideo2} utilizes masked reconstruction, cross-modal contrastive learning, and next-token prediction to enhance the model's perceptiveness, semantic understanding, and reasoning capabilities.

\begin{figure*}[ht]
    \centering
    \includegraphics[width=\linewidth]{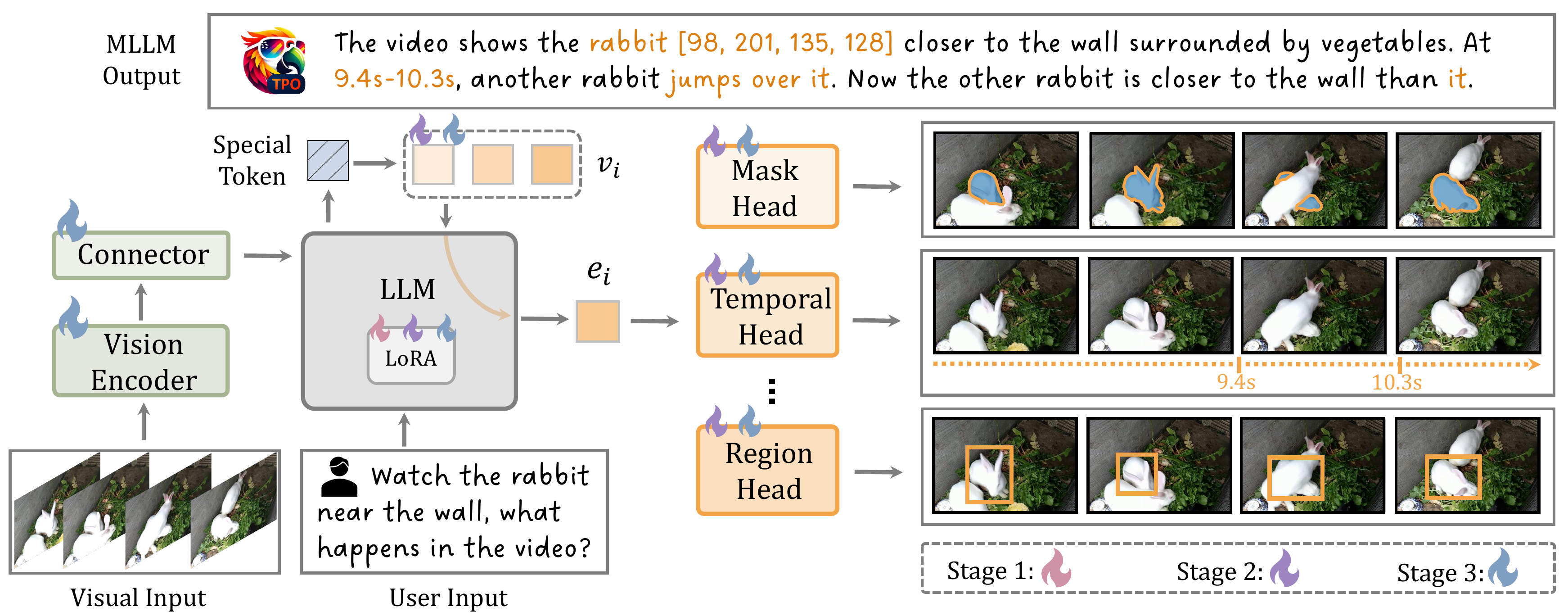}
    \caption{\textbf{Overall Pipeline of TPO.} The architecture of Task Preference Optimization (TPO) consists of four main components: (1) a vision encoder, (2) a connector, (3) a large language model, and (4) a series of visual task heads. Differently colored flame symbols indicate which components are unfrozen at various stages of the training process.}
    \label{fig:overview}
\end{figure*}

Based on vision foundation models, some studies~\cite{lu2022unified, Uni-perceiver, Uni-perceiver_v2, Uni-perceiver-moe, wang2023seggpt} intend to incorporate downstream task heads into this framework and expect end-to-end task training.
Unified-IO~\cite{lu2022unified} introduces a versatile model capable of handling a wide range of tasks across vision and language domains, requiring minimal or no task-specific modifications or additional parameters.
Uni-Perceivers~\cite{Uni-perceiver, Uni-perceiver_v2, Uni-perceiver-moe} formulate different tasks to find the maximum likelihood target for each input through the representation similarity regardless of their modality.
Nevertheless, these generalist models are limited to predefined tasks and cannot support flexible open-ended task customization based on language instructions.

\vspace{-2mm}
\paragraph{Multimodal Large Language Model.}
The effective understanding and reasoning of LMMs has attracted the attention of numerous researchers. Limited to its input modality, researchers expand the visual capabilities of LLM, leading to MLLMs. Seminal works, such as BLIP-2~\cite{pmlr-v202-li23q}, LLaVA~\cite{llava}, and mPLUG-Owl~\cite{mplugowl}, introduce image captioning and visual question answering based on LLM using visual instruction-tuning data. Some video-based MLLMs have been proposed, such as VideoChat~\cite{videochat}, VideoChatGPT~\cite{video-chatgpt}, and Video-LLaMA~\cite{videollama}, enabling LLM to gain video understanding capabilities by encoding multiple video frames and using video instruction data.

Vanilla MLLMs have achieved impressive results in visual-related dialogues, but barely address some fine-grained tasks, such as segmentation and temporal grounding. To address this challenge, MLLMs usually take one of the following pipelines: pixel-to-sequence (P2S) and pixel-to-embedding (P2E).
For P2S methods~\cite{chen2023shikra, Ren2023TimeChat, hawkeye, yu2025merlin, grounded-videollm}, MLLM directly outputs the textual predictions. TimeChat~\cite{Ren2023TimeChat} introduces a time-aware frame encoder and a sliding video Q-former to enhance temporal perception.
For P2E methods~\cite{lai2024lisa, zhang2023nextchat, bai2024one, wang2024visionllm, wu2024visionllm}, MLLM compresses and inputs visuals into the downstream decoder for estimations.
LISA~\cite{lai2024lisa}, NExT-Chat~\cite{zhang2023nextchat}, and VideoLISA~\cite{bai2024one} introduce SAM~\cite{kirillov2023segment} as a segmentation tool, using a special token as a prompt to connect MLLM and SAM.
VisionLLM v2~\cite{wu2024visionllm} designs routing tokens and super-link queries to bridge MLLM with multiple decoders.
\vspace{-2mm}
\paragraph{Alignment in MLLM.}
Aligning MLLM with human preferences or values is crucial for MLLM's development. Recent works~\cite{sun2023aligning, zhou2024aligning, yu2024rlhf} introduce alignment approaches to MLLM, including proximal policy optimization (PPO)~\cite{ppo} and direct preference optimization (DPO)~\cite{rafailov2024direct}, as shown in Figure~\ref{fig:intro}. They usually exploit proprietary models like GPT4-V to build visual preference datasets and then tune MLLMs using PPO or DPO.
Llava-RLHF~\cite{sun2023aligning} incorporates PPO into a MLLM framework Llava, argumented by image captions or question-answers (QA), while Zhou et al.~\cite{zhou2024aligning} and Zhang et al.~\cite{zhang2023llavar} give DPO implementations for MLLMs where visual preference data are created by GPT-4V and other open-sourced MLLMs. RLHF-V~\cite{yu2024rlhf} collects a dense human preference in segmentation and enhances MLLMs with DPO.

\section{Method}
\label{method}

Task Preference Optimization (TPO) aims to enable MLLMs to master classical visual perceptions (such as tracking, temporal grounding, etc.) for better task generalization, regarding that many multimodal reasoning tasks require precise visual cues for accurate and reliable responses. As shown in Figure ~\ref{fig:overview}, MLLM-TPO has a typical multimodal model $M$ (consisting of
a vision encoder $E$, a vision-language connector $C$, and a large language model $G$) and a task preference model $P$ with a series of visual task heads $\{H_i\}_{i=1...n}$ ($n$ denotes the task head number). These heads connect with MLLM using the embeddings $\{\mathbf{e}_i\}_{i=1...n}$ ($\mathbf{e}_i = G(\mathbf{v}_i) \in \mathbb{R}^{1\times \text{C}}$) transformed from the learnable task-specific tokens $\{\mathbf{v}_i\}_{i=1...n}$ ($\mathbf{v}_i \in \mathbb{R}^{1\times \text{C}}$) via MLLM.

TPO employs a local-to-global training scheme, first adapting task heads to the MLLM and then training them jointly. Specifically, the MLLM $M$ starts to recognize $\{\mathbf{v}_i\}_{i=1...n}$ by updating $G$ according to user instructions, then we tune $\{\mathbf{v}_i\}_{i=1...n}$ and $\{H_i\}_{i=1...n}$ for adapting visual heads to $M$. Finally, we train both $M$ and $\{H_i\}_{i=1...n}$ together. During inference, MLLM-TPO can respond to users' queries in text, and produce structured visual outputs (like masks, timestamps, trajectories, etc.) when users ask (\eg yielding time results for ``find when the birthday party starts''). We detail how to form MLLM-TPO structurally and train it in the following.

\subsection{Task Preference Models}
The task preference model (TPM) $P$ contains a series of task tokens $\{\mathbf{v}_i\}_{i=1...n}$ and heads $\{H_i\}_{i=1...n}$. Before TPM works, the attached MLLM generates special tokens indicating task types from the input queries. Then TPM dynamically calls the task token $\mathbf{v}_j$ based on the special token, transforms it to the task embedding $\mathbf{e}_j$ via the LLM $G$ (the last hidden embedding), and feeds $\mathbf{e}_j$ to the corresponding visual task head $\{H_j\}$ for specific task predictions. Considering existing MLLMs demonstrate remarkable capabilities in common object and scene recognition, yet struggle to accurately locate things or actions, the employed task heads mainly focus on spatiotemporal localization and tracking. Specifically, we give three fundamental task types for compensating mainstream MLLMs' gaps from expert models in visual perceptions: 1) region head, 2) temporal head, and 3) mask head. Their architectures are given below.

\begin{itemize}
    \item

\textbf{Region Head.}
A two-layer multilayer perceptron (MLP) with ReLU activations is employed for the region head. It takes embeddings from LLM and regresses them to the bounding box coordinates for spatial grounding.

\item
\textbf{Temporal Head.}
This head is composed of a video encoder, text encoder, and temporal grounding model for moment retrieval and highlight detection. The video and text queries are input into the video and text encoders to obtain their features respectively. Then we append the temporal task embedding after the text features. With the new text and video features, the MLLM estimates the start/end time and highlight score of the corresponding moment from the query via the temporal grounding model.

\item
\textbf{Mask Head.} 
Pixel-level tasks pose significant challenges for MLLMs due to the lack of the corresponding learning data and optimization in MLLMs. To this end, we introduce a specialized mask head, utilizing the image encoder, mask decoder, and memory bank components from SAM2 ~\cite{ravi2024sam2}, replacing the prompt encoder with a single MLP layer called the mask adapter. Specifically, for the given visuals and user queries, the features extracted from the aforementioned image encoder and the text representation corresponding to the mask task embeddings from LLM are employed,  fed into the mask decoder to produce the desired mask.
\end{itemize}

Most known discriminative vision tasks can be addressed by one or a combination of these three task heads. TPM builds the architectural foundations for leveraging existing vision data with annotations to enhance MLLMs.

\subsection{Task Preference Optimization}

TPO improves MLLMs with extra supervisions from visual task heads by back-propagating gradients from heads to update MLLMs using visual task data. It enables the language model $G$ in MLLM to discriminate specific task types when users demand (task assignment). Then, TPO trains TPM $P$ via compact task representations $\mathbf{e}_i$ (task optimization) from $\mathbf{v}_i$. Lastly, we train $M$ and $P$ together, tuning $M$ for the refined spatiotemporal perception according to task preferences from $P$. Its optimization objective is given as:
\begin{equation}
    \mathcal{L} = \mathcal{L}_{\text{mllm}} + \underbrace{\mathcal{L}_{\text{assign}}(G(\mathbf{T}_{\text{q}}), \mathbf{s})}_{\text{Task Assignment}} + \sum_{i=1}^{n} \underbrace{\mathcal{L}_{\text{task}}(\mathbf{A}_i, H_i(G(\mathbf{v}_i)))}_{\text{Task Optimization}},
\end{equation}
where $\mathbf{T}_{\text{q}}$ and $\mathbf{s}$ denote a user query (some contain specific task needs like tracking) in the form of text token sequence and the ground truth task indicating token, respectively. $\mathbf{A}_i$ is the task annotation of the input visual $\mathbf{X}$, so it could be a number tuple for describing regional localization and area or mask for delineating object shape and position. Here $(\mathbf{X}, \mathbf{T}_{\text{q}}, \mathbf{T}_{\text{a}}, \mathbf{A}_i)$ stands for an input data tuple to MLLM-TPO. Usually, we have $(\mathbf{X}, \mathbf{T}_{\text{q}}, \mathbf{T}_{\text{a}})$ for typical MLLM training while $(\mathbf{X}, \mathbf{T}_{\text{q}}, \mathbf{A}_i)$ for task token and head training. $\mathcal{L}_{\text{assign}}$ is the cross-entropy loss. Meanwhile $\mathcal{L}_{\text{task}_i}$ varies according to the task, and is usually regression- or classification-related loss.

To train MLLM-TPO, we propose a 3-stage local-to-global training scheme, given in Figure \ref{fig:overview}. Stage 1 learns to identify the task type based on user queries. In stage 2, we train task heads along with their corresponding task tokens, respectively. Lastly, we co-train task heads with MLLM by both task data and multimodal conversation data. Our 3-stage training strategy mitigates the risk of degrading the MLLM's general abilities. 
We describe them as:
\vspace{-3mm}
\paragraph{Stage 1: Task Assignment.} We design a variety of dialogue templates for different visual tasks to perform task recognition instruction tuning for LLM. The LLM is trained using LoRA~\cite{hu2021lora} in this stage. 

\vspace{-3mm}
\paragraph{Stage 2: Vision Task Training.} We integrate task tokens and task-specific heads into the model. By training on task-specific data, the model equips the capacity with fine-grained perception and aligns between the task head and the MLLM. The region head, temporal head, mask adapter, and corresponding task tokens are trained respectively. Similarly, the LLM is updated with LoRA. 

\vspace{-3mm}
\paragraph{Stage 3: Multi-task Training.} 
We tune the entire model on the mixed corpus combining multimodal conversations and specific task data. The synergistic training allows gradients from the heads to MLLM, supervised by human annotations in vision tasks. The vision encoder, connector, task tokens, region head, temporal head, mask adapter, and LLM (with LoRA) are joint-trained in this stage.

\section{Experiment}
\label{experiment}

\begin{table}[t]
\resizebox{\linewidth}{!}{
    \begin{tabular}{l|c|c}
    \toprule
    \textbf{Task} & \textbf{Samples} & \textbf{Datasets}\\
    \midrule 
    Segmentation & 114.6K & MeViS~\cite{mevis}, SAMv2~\cite{ravi2024sam2}\\
    \midrule 
    \makecell[l]{Temporal \\Grounding}  & 116.5K & \makecell[c]{ActivityNet~\cite{caba2015activitynet}, TACoS~\cite{regneri2013grounding},  QVHighlight~\cite{qvhighlight},\\ DiDeMo~\cite{DiDeMo}, QuerYD~\cite{queryd}, HiREST~\cite{hirest}, NLQ~\cite{grauman2022ego4d}} \\
    \midrule 
    \makecell[l]{Spatial \\Grounding} & 540.0K & \makecell[c]{Allseeing-V2~\cite{wang2024allseeing_v2}, Visual Genome~\cite{krishna2017vg}, \\ RefCOCO~\cite{yu2016refcoco}, RefCOCO+~\cite{yu2016refcoco}, RefCOCOg~\cite{mao2016refcocog}}  \\
    \midrule 
    Conversation &3M & \makecell[c]{YouCook2~\cite{DaXuDoCVPR2013}, ActivityNet~\cite{caba2015activitynet}, \\
    VideoChat2-IT~\cite{mvbench}, ShareGPT-4o~\cite{chen2023internvl}, \\ 
    LLaVA-Hound-DPO~\cite{zhang2024direct},  ShareGPT4V~\cite{chen2023sharegpt4v} } \\
    \bottomrule 
\end{tabular}
}
\vspace{-3mm}
\caption{\textbf{Overview of Datasets Used in TPO for Various Tasks.}}
\label{tab:dataset}
\end{table}

We give the implementation and training \& testing specifics of our TPO, and then show its results and ablations.
\vspace{-2mm}
\paragraph{Implementation Details.}
We employ VideoChat2~\cite{mvbench}, a video-based MLLM, as the primary framework in experiments.
For its configuration, we employ UMT-L~\cite{li2023unmasked} as the vision encoder, Q-former in BERT$_{\text{base}}$~\cite{BERT} as the connector, and Mistral-7B~\cite{mistral} as the language model (LLM).

Regarding the three task heads, the Region Head is initialized randomly.  The Temporal Head utilizes CG-DETR~\cite{moon2023correlation}, with parameters also initialized randomly. The video features input to the temporal head are extracted from the pre-trained InternVideo2~\cite{wang2024internvideo2}, while query features are extracted using the LLM~\cite{chinese-llama-alpaca}.
The Mask Head employs SAM2~\cite{ravi2024sam2} and is initialized with its pre-trained weights. Since SAM2 is originally designed solely for tracking, we incorporate a two-layer MLP to encode positioning prompts for spatiotemporal grounding. Additionally, to enable the MLLM to handle the spatial locations provided by the user, we utilize another two-layer MLP to encode the spatial input into the MLLM.

Our model is trained on a variety of visual task datasets and conversation datasets, as shown in Table~\ref{tab:dataset}.

More training details are provided in the Appendix.
\vspace{-2mm}
\paragraph{Benchmarks.} We evaluate TPO on both general image/video understanding benchmarks (mainly in multiple choice form) and specific visual tasks (e.g. grounding, tracking, and so on). Specifically, we run our model along with other approaches on MVBench, VideoMME, NExT-GQA, MLVU, MMIU, and SEED-Bench2. \textbf{MVBench}~\cite{mvbench} is designed to evaluate multi-modal fine-grained video understanding tasks for clips (lasting around 8 to 16 seconds), consisting of 20 video tasks relying heavily on temporal perception that can hardly be addressed by single-frame analysis. \textbf{Video-MME}~\cite{videomme} is for evaluating MLLMs in both perception and reasoning across varying lengths of videos, annotated by humans. \textbf{NExT-GQA}~\cite{nextgqa} builds on NExT-QA by introducing timestamps that are crucial for understanding questions and determining answers. This requires MLLM to perform multi-step reasoning and emphasize a deeper understanding of both visual and textual content.
\textbf{MLVU}~\cite{MLVU} is constructed from a wide variety of long videos, with lengths ranging from 3 minutes to 2 hours, and includes nine distinct evaluation tasks.
\textbf{MMIU}~\cite{meng2024mmiu} is a comprehensive evaluation suite designed to assess MLLMs across a wide range of multi-image tasks. It encompasses 7 types of multi-image relationships, 52 tasks, 77K images, and 11K meticulously curated multiple-choice questions.
\textbf{SEED2-Bench2}~\cite{seedbench2} is a  MLLM benchmark, featuring 24K multiple-choice questions with human annotations. It spans 27 evaluation dimensions, assessing both text and image generation.

Concerning visual tasks, we test spatial grounding, moment retrieval, highlight detection, tracking, and referring segmentation, including 7 related benchmarks ~\cite{charades-sta, qvhighlight, yu2016refcoco, fan2019lasot, huang2019got, mevis, seo2020urvos} ,  and the results from several corresponding state-of-the-art expert models.

\subsection{General Understanding Evaluation}

\begin{table*}[ht]
\centering
\resizebox{\linewidth}{!}{%
\begin{tabular}{l|c|c|c|cccccccc|c}
\toprule
\multirow{3}{*}{Model} & \multirow{3}{*}{\begin{tabular}[c]{@{}c@{}}LLM\\ Params\end{tabular}} & \multirow{3}{*}{Frames} & \multirow{2}{*}{MVBench~\cite{mvbench}} & \multicolumn{8}{c|}{VideoMME~\cite{videomme}} & \multirow{2}{*}{MLVU~\cite{MLVU}} \\ \cmidrule(lr){5-6} \cmidrule(lr){7-8} \cmidrule(lr){9-10} \cmidrule(lr){11-12}
 &  &  &  & \multicolumn{2}{c|}{Overall} & \multicolumn{2}{c|}{Short} & \multicolumn{2}{c|}{Medium} & \multicolumn{2}{c|}{Long} &    \\ \cmidrule(lr){4-4} \cmidrule(lr){5-6} \cmidrule(lr){7-8} \cmidrule(lr){9-10} \cmidrule(lr){11-12}  \cmidrule(lr){13-13}
 &  &  & AVG & w/o s. & w/ s. & w/o s. & w/ s. & w/o s. & w/ s. & w/o s. & w/ s. & M-AVG \\ 
\midrule
TimeChat~\cite{Ren2023TimeChat} & 7B & 96 & 38.5 & 34.3 & 36.9 & 39.1 & 43.1 & 31.8 & 33.9 & 32.1 & 33.6 & 30.9 \\
Video-LLAVA~\cite{video-llava} & 7B & 8 & 43.0 & 41.1 & 41.9 & 46.9 & 47.3 & 38.7 & 40.4 & 37.8 & 37.9 & 47.3 \\
ShareGPT4Video~\cite{sharegpt4video} & 7B & 16 & 51.2 & 39.9 & 43.6 & 48.3 & 53.6 & 36.3 & 39.3 & 35.0 & 37.9 & 46.4 \\
LLaVA-Next-Video~\cite{llavanext-video} & 7B & 16 & 44.0 & 38.0 & 40.8 & 44.6 & 47.4 & 37.7 & 39.4 & 31.9 & 35.6 & 39.3 \\
ST-LLM~\cite{st-llm} & 7B & 64 & 54.9 & 37.9 & 42.3 & 45.7 & 48.4 & 36.8 & 41.4 & 31.3 & 36.9 & - \\
PLLaVA-34B~\cite{pllava} & 34B & 16 & 58.1 & 40.0 & 35.0 & 47.2 &  36.2 &  38.2 &  35.9 & 34.7 & 32.9 & 53.6 \\
Chat-UniVi~\cite{jin2023chatunivi} & 7B & 64 & 40.8 & 40.6 & 45.9 & 45.7 & 51.2 & 41.3 & 47.3 &39.1 & 43.4 & - \\
VideoChat2(baseline)~\cite{mvbench} & 7B & 16 & 60.4 & 39.5 & 43.8 & 48.3 & 52.8 & 37.0 & 39.4 & 33.2 & 39.2 & 44.5 \\
\midrule
VideoChat-TPO & 7B & 16 & \textbf{66.8}~\textcolor{olive}{(+6.4)} & \textbf{48.8}~\textcolor{olive}{(+9.3)} & \textbf{53.8}~\textcolor{olive}{(+10.0)} & \textbf{58.8} & \textbf{64.9} & \textbf{46.7} & \textbf{50.0} & \textbf{41.0} & \textbf{46.4} & \textbf{54.7}~\textcolor{olive}{(+10.2)} \\ 
\bottomrule
\end{tabular}
}
\vspace{-3mm}
\caption{\textbf{Performance on Multimodal Video Understanding.}
We compare our model to others using LLMs of the same generation or 16-frame input. w/o s. indicates without subtitle, while w s. indicates with subtitle. M-AVG refers to the mean average of MLVU.}
\label{tab:performance_comparison}
\end{table*}

\paragraph{Multimodal Video Understanding.} TPO improves its baseline (VideoChat2) on several video understanding benchmarks with a notable margin.
As shown in Table~\ref{tab:performance_comparison}, VideoChat-TPO, using a 7B LLM and only 16 input frames, achieves a 66.8 average score on MVBench~\cite{mvbench}, increasing by 6.4\% over VideoChat2 and exceeding the performance of ST-LLM~\cite{st-llm} which uses 64 frames. Considering MVBench focuses on characterizing subtle temporal variations involving fine-grained action types, action order, moving direction, and so on, TPO's seamless integration of detailed video understanding (like spatiotemporal grounding and tracking) into its optimized MLLM makes it effectively handle these fine-grained tasks.

On VideoMME~\cite{videomme}, VideoChat-TPO outperforms the compared models, achieving a 9.3\% improvement over VideoChat2 and demonstrating significant gains across short and medium-length videos, regardless of subtitle availability. On MLVU~\cite{MLVU}, VideoChat-TPO achieves a M-AVG score of 54.7, surpassing VideoChat2 by 10.2\%. 
This confirms the effectiveness of TPO's enhanced perceptions of long-form video evaluations.
These results across three benchmarks demonstrate the notable advancements in multimodal video understanding achieved through TPO. 

\begin{table*}[ht]
\centering
\begin{minipage}{\linewidth}
    \centering
    \begin{minipage}{0.6\linewidth}
        \centering
        \resizebox{\linewidth}{!}{
        \begin{tabular}{l|cc|cccccc}
        \toprule
        Model &Acc@IoP &Acc@GQA &mIoP &IoP@0.3 &IoP@0.5 &mIoU &IoU@0.3 &IoU@0.5 \\
        \midrule 
        VIOLETv2 \cite{violetv2}         & 54.9 &12.8 & 23.6 & 25.1 & 23.3 & 3.1  & 4.3  & 1.3 \\  
        SeViLA \cite{sevila}             & 72.5 &16.6 & 29.5 & 34.7 & 22.9 & 21.7 & 29.2 & 13.8\\
        LangRepo \cite{langrepo}         & 59.6 &17.1 & 31.3 & -    & 28.7 & 18.5 & -    & 12.2 \\ 
        FrozenBiLM NG+ \cite{frozenbilm} & 73.8 &17.5 & 24.2 & 28.5 & 23.7 & 9.6  & 13.5 & 6.1 \\  
        VideoStreaming \cite{streaming}  & 57.4 &17.8 & 32.2 & -    & 31.0 & 19.3 & -    & 13.3 \\  
        LLoVi  \cite{llovi}              & 65.9 &24.3 & \textbf{37.3} & -    & \textbf{36.9} & 20.0 & -  & 15.3\\  
        HawkEye \cite{hawkeye}           & -    &- & -    & -    & -    & 25.7 & 37.0 & 19.5 \\  
        \midrule 
        VideoChat-TPO       & \textbf{77.7} & \textbf{25.5} & 35.6 & \textbf{47.5} & 32.8 & \textbf{27.7} & \textbf{41.2} & \textbf{23.4} \\  
        \bottomrule 
        \end{tabular}
        }
        \vspace{-3mm}
        \caption{\textbf{Performance on Grounded QA.} }
        \label{tab:results_nextgqa}
    \end{minipage}%
    \begin{minipage}{0.4\linewidth}
        \centering
        \resizebox{\linewidth}{!}{
        \begin{tabular}{l|c|c|c}
        \toprule
        Model & MMIU~\cite{meng2024mmiu} & SEED2$_{I}$~\cite{seedbench2} & SEED2$_{M}$~\cite{seedbench2} \\
        \midrule 
        LLaVA-v1.5~\cite{llava}  & 19.2  & 58.3  & 39.2  \\  
        ShareGPT4V~\cite{chen2023sharegpt4v} & 18.5 & - & - \\
        OpenFlamingo~\cite{awadalla2023openflamingo} & 22.3 & 36.6  & 43.5 \\
        LLaVA-Interleave~\cite{li2024llava}  & 32.4 & -  & - \\
        VideoChat2~\cite{mvbench}    & 35.0 & 26.5  & 27.6 \\ 
        VideoChatGPT~\cite{video-chatgpt} & - & 38.3 & 49.8 \\
        InternLM-XComposer~\cite{ixc2} & 21.9 & 65.4  & 49.8 \\
        \midrule  
        VideoChat-TPO  &\textbf{40.2}~\textcolor{olive}{(+5.2)}   & \textbf{67.3}~\textcolor{olive}{(+40.8)}  &  \textbf{70.0}~\textcolor{olive}{(+42.4)}  \\ 
        \bottomrule
        \end{tabular}
        }
        \vspace{-3mm}
        \caption{\textbf{Performance on Image Understanding.} }
        \label{tab:MMIU}
    \end{minipage}
\end{minipage}
\end{table*}

\vspace{-3mm}
\paragraph{Grounded Video QA.} Table~\ref{tab:results_nextgqa} shows that VideoChat-TPO outperforms other models, achieving superior accuracy (Acc) and intersection over union (IoU) scores in NExT-GQA~\cite{nextgqa}. Its intersection over prediction (IoP) scores are comparable to those of LLoVi \cite{llovi}, which employs large, closed-source commercial models like GPT-4~\cite{gpt4}. 
The high Acc@IoP score not only reflects the enhanced capability of TPO-optimized VideoChat in effectively understanding and interpreting video content, but also demonstrates its robustness in handling complex reasoning scenarios. Furthermore, the higher Acc@GQA score indicates that the model successfully integrates fine-grained temporal grounding with complex reasoning tasks, enabling it to accurately provide temporal clues necessary for inferring answers.
From this perspective, TPO training framework significantly allows MLLMs to overcome their limitations in identifying and localizing temporal elements within video data. This enhancement positions MLLM-TPO with a competitive edge in temporal-related reasoning.

\vspace{-3mm}
\paragraph{Multimodal Multi-image Understanding.} To explore TPO's effect on multi-image understanding, we test it on the MMIU and SEED-Bench2, as shown in Table~\ref{tab:MMIU}. On MMIU, VideoChat-TPO achieved an overall score of 40.2, a 5.2\% improvement over VideoChat2. Besides temporal changes, MMIU evaluates models' spatial sensing and semantic relations in scenes.
On SEED-Bench2, across 27 performance indicators, VideoChat-TPO achieves 41.7\% improvement on average performance compared to VideoChat2. 
VideoChat-TPO's superior performance on multi-image understanding compared to LLaVA-Interleave and InternLM-XComposer demonstrates that TPO's vision enhancements improve the MLLM's spatial perception and image understanding.

\subsection{Vision Task Evaluation}
\paragraph{Moment Retrieval.} 
Moment Retrieval is to locate the target segments in a video based on queries. 
Table \ref{tab:tvg_zs} and \ref{tab:tvg_ft} compare the zero-shot and fine-tuned moment retrieval performance of \ModelName{} against other expert models and MLLMs. In zero-shot settings, \ModelName{} achieves a R@1 (IoU=0.5) score of 40.2 on the Charades-STA \cite{charades-sta}. It surpasses the previous state-of-the-art (SOTA) MLLM specialized for temporal tasks, ChatVTG \cite{chatvtg}, and the expert model, UniVTG \cite{univtg}.
This demonstrates \ModelName{}'s ability to accurately locate video moments corresponding to given text queries, thereby enhancing the practical applicability of MLLMs.
\vspace{-2mm}
\paragraph{Highlight Detection.}

Highlight detection generates salient scores for emphasizing frames based on the input language query. We compare the fine-tuning highlight detection performance of \ModelName{} with that of other expert models and MLLMs, as shown in Table~\ref{tab:tvg_ft}. \ModelName{} notably outperforms the previous state-of-the-art MLLM (such as TimeChat~\cite{Ren2023TimeChat}) on both Charades-STA and QVHighlight. For expert models like QD-DETR and UniVTG, \ModelName{} beats them on Charades-STA non-trivially and achieves a comparable performance on QVHighlight. This also demonstrates the progress in extending MLLMs to broad temporal tasks.
\begin{table*}[ht]
\centering
\begin{minipage}{\linewidth}
    \centering
    \begin{minipage}{0.3\linewidth}
          \centering
            \resizebox{\linewidth}{!}{
            \begin{tabular}{l|ccc|c}
            \toprule
            \multirow{2}{*}{Model} &\multicolumn{4}{c}{Charades-STA~\cite{charades-sta}} \\
            \cmidrule(lr){2-5} &R@0.3 &R@0.5 &R@0.7 &mIoU\\
            \midrule 
            \textcolor{gray}{UniVTG~\cite{univtg}} & \textcolor{gray}{44.1} & \textcolor{gray}{25.2}  & \textcolor{gray}{10.0} & \textcolor{gray}{27.1} \\
            \midrule 
            VideoChat2~\cite{mvbench}& 38.0 & 14.3 & 3.8  & 24.6\\
            VTimeLLM~\cite{vtimellm}  & 51.0 & 27.5 & 11.4 & 31.2\\
            TimeChat~\cite{Ren2023TimeChat}  & -    & 32.2 & 13.4 & -   \\
            HawkEye~\cite{hawkeye}   & 50.6 & 31.4 & 14.5 & 33.7 \\
            ChatVTG ~\cite{chatvtg}         &52.7 & 33.0 & 15.9 &34.9 \\      
            \midrule 
            \ModelName   & \textbf{58.3} & \textbf{40.2} & \textbf{18.4} & \textbf{38.1}\\  
            \bottomrule 
        \end{tabular}}
         \vspace{-3mm}
        \caption{\textbf{Zero-Shot Performance on Moment Retrieval.}\textcolor{gray}{Gray means no LLM.} }
        \label{tab:tvg_zs}
    \end{minipage}%
    \begin{minipage}{0.44\linewidth}
              \centering
                \resizebox{\linewidth}{!}{
                \begin{tabular}{l|ccc|c|cc}
                \toprule
                \multirow{2}{*}{Model} &\multicolumn{4}{c}{Charades-STA~\cite{charades-sta}} &\multicolumn{2}{|c}{QVHighlight~\cite{qvhighlight}} \\
                \cmidrule(lr){2-5} \cmidrule(lr){6-7}
                &R@0.3 &R@0.5 &R@0.7 &mIoU &mAP &HIT@1\\
                \midrule 
                \textcolor{gray}{M-DETR~\cite{qvhighlight}  }  & \textcolor{gray}{65.8} & \textcolor{gray}{52.1}  & \textcolor{gray}{30.6} & \textcolor{gray}{45.5} & \textcolor{gray}{35.7}  & \textcolor{gray}{55.6}  \\
                \textcolor{gray}{QD-DETR~\cite{qd-detr}}  & \textcolor{gray}{-}    & \textcolor{gray}{57.3 } & \textcolor{gray}{32.6} & \textcolor{gray}{-}    & \textcolor{gray}{38.9}  & \textcolor{gray}{62.4} \\
                \textcolor{gray}{UniVTG~\cite{univtg}}    & \textcolor{gray}{72.6} & \textcolor{gray}{60.2}  & \textcolor{gray}{38.6} & \textcolor{gray}{52.2} & \textcolor{gray}{\textbf{40.5}}  & \textcolor{gray}{\textbf{66.3}} \\
                \midrule
                TimeChat~\cite{Ren2023TimeChat}& - & 46.7 & 23.7 & -    & 21.7 & 37.9  \\
                HawkEye~\cite{hawkeye}      & 72.5 & 58.3 & 28.8 & - & - & -\\
                \midrule 
                 \ModelName   & \textbf{77.0} & \textbf{65.0} & \textbf{40.7} & \textbf{55.0} & \underline{38.6} & \underline{65.4}\\  
                \bottomrule 
            \end{tabular}
            \captionsetup{ width=0.95\linewidth}}
         \vspace{-3mm}
         \captionsetup{ width=0.95\linewidth}
         \caption{\textbf{Fine-tuning Performance on Moment Retrieval and Highlight Detection.}~\textcolor{gray}{Gray means no LLM.}}
        \label{tab:tvg_ft}
    \end{minipage}
    \begin{minipage}{0.24\linewidth}
            \centering
            \resizebox{\linewidth}{!}{
            \begin{tabular}{l |c c c}
            \toprule
            \multirow{2}{*}{Methods} & \multicolumn{3}{c}{RefCOCO~\cite{yu2016refcoco}}  \\
             \cmidrule(lr){2-4}  
             & val & testA & testB \\
            \midrule
            \textcolor{gray}{MAttNet $\bigstar$~\citep{yu2018mattnet}} & \textcolor{gray}{76.4} & \textcolor{gray}{80.4} & \textcolor{gray}{69.3} \\
            OFA-L~\citep{wang2022ofa} & 80.0 & 83.7 & 76.4 \\
            \textcolor{gray}{G-DINO-L $\bigstar$~\citep{liu2023gdino}} & \textcolor{gray}{\textbf{90.6}} & \textcolor{gray}{\textbf{93.2}} & \textcolor{gray}{\textbf{88.2}} \\
            \midrule
            VisionLLM-H~\citep{wang2024visionllm} & - & 86.7 & - \\
            Shikra-7B~\citep{chen2023shikra} & \textbf{87.0} & 90.6 & 80.2  \\
            NExT-Chat-7B~\cite{zhang2023nextchat}  & 85.5 & 90.0 & 77.9  \\
            \midrule
             VideoChat-TPO & 85.9 & \textbf{90.8} & \textbf{81.3} \\
            \bottomrule
            \end{tabular}}
            \vspace{-3mm}
            \caption{\textbf{Spatial Grounding Task.} 
            \textcolor{gray}{$\bigstar$ with a refined decoder.}}
            \label{tab:rec}
    \end{minipage}
\end{minipage}
\end{table*}
\vspace{-2mm}
\paragraph{Spatial Grounding.}
    To verify the fine-grained localization ability of the model, we run the spatial grounding task which inputs textual descriptions into the model and gets the corresponding bounding box on RefCOCO~\cite{yu2016refcoco}. As shown in Table~\ref{tab:rec}, we compare VideoChat-TPO with pixel-to-sequence models, \ie, VisionLLM-H~\cite{wang2024visionllm},  pixel-to-embedding methods, \ie, NExT-Chat~\cite{zhang2023nextchat}, and expert models, \ie, G-DINO~\cite{liu2023gdino}. VideoChat-TPO performs better than the pixel-to-embedding methods by using only a simple task head and achieves comparable performance to pixel-to-sequence models fine-tuned on a large amount of spatial grounding data, as well as to specialized models.

\begin{table*}[ht]
\centering
\begin{minipage}{\linewidth}
    \begin{minipage}{0.38\linewidth}
        \centering
        \resizebox{\columnwidth}{!}{
        \begin{tabular}{l|ccc|ccc}
        \toprule
        \multirow{2}{*}{Model} & \multicolumn{3}{c|}{LaSOT~\cite{fan2019lasot}} & \multicolumn{3}{c}{GOT-10k~\cite{huang2019got}} \\
        \cmidrule{2-4} \cmidrule{5-7}    
        & Success & P$_{norm}$ & P & Overlap & SR0.5 & SR0.75\\
        \midrule
        SiamFC~\cite{siamFC}              & 33.6 & 42.0 & 33.9  & 34.8 & 35.3 & 9.8 \\
        ATOM~\cite{ATOM}                 &   51.5 & -  &-  & 55.6 & 63.4 & 40.2\\
        SiamRPN++~\cite{SiamRPN++}      &  49.6 & 56.9 & 49.1 & 51.8 & 61.8 & 32.5\\
        SiamFC++~\cite{SiamFC++}          &  54.4 & 62.3 & 54.7 & 59.5 & 69.5 & 47.9\\
        \midrule    
        LLaVA-1.5~\cite{llava}  & 19.4 & 16.5 & 12.8 & 23.5 & 20.2 & 9.7 \\
        Merlin~\cite{yu2025merlin}  & 39.8   & 40.2  & 38.1 & 51.4 & 55.9 & 42.8\\
        \midrule
        VideoChat-TPO  & \textbf{69.4}  & \textbf{80.1}  & \textbf{76.9} & \textbf{70.6} & \textbf{79.8} & \textbf{66.0}\\
        \bottomrule
        \end{tabular}
        }
        \vspace{-3mm}
        \caption{\textbf{Performance on Tracking Benchmarks.}}
        \label{tab:track}
    \end{minipage}
    \begin{minipage}{0.38\linewidth}
        \centering
        \resizebox{\linewidth}{!}{
        {
            \begin{tabular}{l | c c c | c c c}
            \toprule
            \multirow{2}*{Method} &  \multicolumn{3}{c |}{Ref-YouTube-VOS ~\cite{seo2020urvos}} & \multicolumn{3}{c}{MeViS~\cite{mevis}} \\
              & \( \mathcal{J} \)\&\( \mathcal{F} \) & \( \mathcal{J} \) & \( \mathcal{F} \)  &  \( \mathcal{J} \)\&\( \mathcal{F} \) & \( \mathcal{J} \) & \( \mathcal{F} \) \\
            \midrule
            ReferFormer~\cite{wu2022language}  & 62.9 & 61.3 & 64.6 & 31.0 & 29.8 & 32.2 \\
            OnlineRefer~\cite{wu2023onlinerefer}  &62.9& 61.0 &64.7 &- &- &-\\
            \midrule
            LISA~\cite{lai2024lisa} & 52.6 & 52.1 & 53.0 & - & - & -\\
            VideoLISA~\cite{bai2024one}  &63.7 & ~\textbf{61.7} &63.7 &44.4 &41.3 &47.6\\
            \midrule
            {VideoChat-TPO} & ~\textbf{63.9} &52.3 & ~\textbf{75.4} & ~\textbf{47.0} & ~\textbf{42.6} & ~\textbf{51.3} \\
            \bottomrule 
            \end{tabular}
        }
    }
        \vspace{-2mm}
        \caption{\textbf{Performance on Referring Segmentation.} TPO is evaluated in a zero-shot on Ref-YouTube.}
        \label{tab:refvos}
    \end{minipage}
    \begin{minipage}{0.23\linewidth}
        \centering
        \resizebox{\columnwidth}{!}{
        \begin{tabular}{l|cc}
        \toprule
        Model & MVBench \\
        \midrule
        LLaVA-OV~\cite{Llava-onevision} & 56.7 \\
        LLaVA-OV-TPO & \textbf{64.8}~\textcolor{olive}{(+8.1)} \\
        \bottomrule
        \end{tabular}
    }
    \vspace{+0.05mm}
    \caption{\textbf{Applying TPO on LLaVA-OV.} TPO significantly enhances the ability to understand fine-grained video details.}
    \label{tab:ablate_mvbench}
\end{minipage}
\end{minipage}
\end{table*}

\vspace{-2mm}
\paragraph{Tracking.}
In the tracking task, the model receives the object coordinates from the first frame and outputs the coordinates for the remaining frames in the video. We evaluate the \ModelName{} on the mainstream tracking benchmarks LAOST~\cite{fan2019lasot} and GOT-10k~\cite{huang2019got}, as shown in Table \ref{tab:track}. 
In zero-shot testing, \ModelName{} ranks best among all MLLMs, even surpassing some fine-tuned expert models notably with around 10\% increase on success rate, such as SiamFC~\cite{siamFC} and ATOM~\cite{ATOM}. With the TPO training method, the model is optimized from mask sequences that hard to represent by words, allowing the MLLM to achieve strong motion characterization for multimodal generalization in highly dynamic and occluded scenes.

\vspace{-3mm}
\paragraph{Referring Segmentation.}
The referring segmentation task requires the model to output the mask sequence corresponding to the specified prompt.
This capability is newly activated in VideoChat-TPO, as it is not supported or designed in the original VideoChat2 or SAM2.
We compare the referring segmentation performance of VideoChat-TPO with other expert models and MLLMs in Table~\ref{tab:refvos}. The zero shot capability of VideoChat-TPO is comparable to the fine-tuning capabilities of other expert models, \ie ReferFormer ~\cite{wu2022language}, showing its notable open-world video segmentation performance.
Through TPO training, MLLM effectively optimizes its understanding in both tracking and segmenting objects indicated by users. It further enables the model to excel in pixel-level tasks, offering perceptual advantages to it in practical applications like robotic control.

\vspace{-1mm}
\subsection{Ablation Studies}
\label{ablation}

\begin{table}[ht]
  \centering
    \resizebox{\linewidth}{!}{
    \begin{tabular}{l|ccc|c|c}
    \toprule
    \multirow{2}{*}{Model} &\multicolumn{4}{c}{Charades-STA~\cite{charades-sta}} &\multicolumn{1}{|c}{MVBench} \\
    \cmidrule(lr){2-5} \cmidrule(lr){6-6}
    &R@0.3 &R@0.5 &R@0.7 &mIoU & AVG\\
    \midrule
        \ModelName   & \textbf{58.3} & \textbf{40.2} & \textbf{18.4} & \textbf{38.1} & \textbf{66.8}\\  

     ~~~~w/o reasoning data  &  56.4 & 38.3 & 17.1 & 35.6 & 66.1\\  
     ~~~~replace by simple head  & 31.5  & 17.8  & 6.1 &  15.4 & 65.8 \\
     ~~~~textualized task data  & 33.7  & 18.6  & 6.2 &  16.0 & 64.7 \\
    
    \bottomrule 
\end{tabular}
}
\vspace{-3mm}
\caption{\textbf{Ablation of Reasoning Data and Head Performance.}}
\label{tab:ablate_tvg}
\end{table}

\begin{table}[ht]
  \centering
    \resizebox{0.85\linewidth}{!}{
    \begin{tabular}{c|cccc|ccc}
    \toprule
    
    \multirow{1}{*}{Exp.} &\multicolumn{1}{c}{T} &\multicolumn{1}{c}{R} &\multicolumn{1}{c}{M} &\multicolumn{1}{c|}{C} &\multicolumn{1}{c}{R@0.5} &\multicolumn{1}{c}{Acc@0.5} &\multicolumn{1}{c}{\( \mathcal{J} \)\&\( \mathcal{F} \)}\\
    \midrule
     1  & \checkmark &&& &  30.2 &  - & - \\
     2  & & \checkmark && &  - & 77.3 & - \\  
     3  & & & \checkmark & &  - &  - & 55.1 \\  
     4  & &\checkmark & \checkmark& &  - &  80.2 & 58.3 \\  
     5  &\checkmark &\checkmark &\checkmark& &  36.7 &  81.6 & 61.4 \\  
    \midrule
     6  &\checkmark &\checkmark&\checkmark&\checkmark & \textbf{40.2} & \textbf{82.0} & \textbf{63.9} \\  
    
    \bottomrule 
\end{tabular}
}
\vspace{-3mm}
\caption{\textbf{Impact of TPO Components and Data.} T, R, M, and C denote temporal head, region head, mask head, and conversation data respectively. R1@0.5 means R1@0.5 in Charades-STA, Acc@0.5 represents the mean of Acc@0.5 in all COCO datasets, \( \mathcal{J} \)\&\( \mathcal{F} \) means \( \mathcal{J} \)\&\( \mathcal{F} \) in Ref-YouTube-VOS.}
\vspace{-3mm}
\label{tab:ablate_training}
\end{table}

In this section, we analyze the effectiveness of the key components of TPO and evaluate its scalability.

\vspace{-3mm}
\paragraph{Extending to Other MLLMs.}
To evaluate the effectiveness of the TPO method on different MLLMs, we apply TPO to LLaVA (LLaVA-OneVision~\cite{Llava-onevision}) in addition to VideoChat. 
In LLaVA, fine-grained video understanding is a weak spot. Despite the conversation data used in LLaVA-OneVision being very similar to, or even more extensive than, what we employ, there remains significant room for improvement in its performance on short-term fine-grained understanding tasks.
As shown in Table~\ref{tab:ablate_mvbench}, with TPO, LLaVA not only gains capabilities that it originally lacked but also achieves an 8.1\% improvement on MVBench, demonstrating the universality of the TPO method.

\vspace{-3mm}
\paragraph{Task Preference Model vs. Textualized Task Data.}

We compare the TPO method with the approach of textualizing task data and training it in an autoregressive manner, which is a straightforward and efficient method for MLLMs to learn specific tasks.
We use the same data as the TPO method and convert the data related to the task head into conversation format. Due to the limitations of textual representation, much data must be input in a more discrete manner. For instance, in tracking supervision data, we follow Merlin~\cite{yu2025merlin} by converting masks into a sequence of spatial coordinates to serve as both the model's input and output. 

As shown in Table~\ref{tab:ablate_tvg}, MLLM-TPO achieves a 2.1\% performance improvement on MVBench~\cite{mvbench} compared to MLLMs trained with textualized task data. 

This demonstrates that the task head in the TPO method enables more effective utilization of the original supervision signals compared to the next-token prediction approach, which inevitably incurs information loss when converting fine-grained tasks into textual annotations. The next-token prediction method cannot truly capture the nuanced information present in videos, while TPO allows the MLLM to learn perceptual information beyond conversational data.

\vspace{-3mm}
\paragraph{Impact of Task Head Performance on TPO.} A stronger task head is expected to enhance the multimodal capabilities of TPO more effectively than a weaker one. 
To further explore the impact of the task head on model performance, we replaced the temporal head from CG-DETR~\cite{moon2023correlation} with a simple two-layer MLP.

As shown in Table~\ref{tab:ablate_tvg}, the simple temporal head exhibits a significant performance decline compared to CG-DETR on the corresponding tasks. However, it still achieves a 1.1\%  improvement compared to the model without TPO. In contrast to the simpler head, the well-designed head can provide more accurate expert knowledge through its architecture and pre-trained task preference weights, enabling better utilization of the data.

\vspace{-3mm}
\paragraph{Impact of Data Scaling.}
Concerning foundation or large models, we expect to enhance their model performance by incorporating additional data at scale. In our approach, we expand the existing conversation data by integrating two reasoning datasets~\cite{caba2015activitynet, DaXuDoCVPR2013}
with temporal information.
By incorporating these datasets, we observe a notable enhancement in the capabilities of the temporal head in Table~\ref{tab:ablate_tvg}. 
Furthermore, as shown in Table~\ref{tab:ablate_training}, adding conversation data further improves performance across various tasks.
Overall, increasing the amount of conversation and fine-grained task data leads to noticeable improvements in model performance, demonstrating the effectiveness of data scaling in training multimodal models with TPO.

\vspace{-3mm}
\paragraph{Synergistic Gains from Co-training.}
Table~\ref{tab:ablate_training} presents the impact of incorporating different visual tasks into TPO on performance. 
 
This indicates that the collaborative learning of visual tasks facilitates the transfer of knowledge, ultimately leading to better performance across the board. Overall, these results confirm that co-training not only benefits individual tasks but also creates a synergistic effect that enhances overall capabilities in visual understanding.






\section{Conclusions}
\vspace{-1mm}

This paper introduces Task Preference Optimization (TPO). It enhances the overall multimodal performance of MLLMs by enhancing their precise visual understanding. TPO achieves this by integrating differentiable task preferences derived from fine-grained visual tasks. It introduces learnable task tokens that serve as bridges between multiple task-specific heads and the MLLM. Through the joint optimization of these task tokens, heads, and the MLLM, TPO leads to a substantial improvement in multimodal dialogue capabilities and performance across various visual tasks. Our results demonstrate the effectiveness of TPO in scaling MLLMs with task-specific data and seamlessly integrating them with existing expert vision models. We believe this study clarifies the prerequisites for fusing MLLMs with models and data from the pre-large model era, effectively bridging the gap between expert and generalist models, as well as between generation and understanding.

\vspace{-3mm}
\paragraph{Limitations.}Currently, TPO focuses exclusively on discriminative visual tasks, overlooking generative ones. Meanwhile, the framework is supervised by human annotations, which neglects potentially valuable unsupervised or self-supervised learning approaches such as contrastive learning~\cite{chen2020simple, he2019moco}.
This limitation inherently restricts the scalability of TPO in terms of both task diversity and requirements. While we demonstrate TPO's potential for enhancing MLLMs through increased task-specific data, a comprehensive investigation into the broader impact of this scaling remains a crucial area for future research.

\section*{Acknowledgements}
This work is partially supported by the National Key R\&D Program of China (No. 2022ZD0160102), Jiangsu Frontier Technology  R\&D Program (No. BF2024076), and the Science and Technology Commission of Shanghai Municipality under Grant No. 23YF1461900.
{
    \small
    \bibliographystyle{ieeenat_fullname}
    \bibliography{main}
}

\section*{Appendix}
\setcounter{section}{0}

\section{Experiment Details}
\label{sec:rationale}

\begin{table*}[ht]
  \centering
    \resizebox{\linewidth}{!}{
    \begin{tabular}{l|c|cccccccccccccccccccc}
    \toprule
    Model &Avg. &AS &AP &AA &FA &UA &OE &OI &OS &MD &AL &ST &AC &MC &MA &SC &FP &CO &EN &ER &CI \\
    \midrule 
    VideoChatGPT \citep{video-chatgpt} & 32.7 & 23.5 & 26.0 & 62.0 & 22.5 & 26.5 & 54.0 & 28.0 & 40.0 & 23.0 & 20.0 & 31.0 & 30.5 & 25.5 & 39.5 & 48.5 & 29.0 & 33.0 & 29.5 & 26.0 & 35.5 \\
    VideoLLaMA \citep{video-llama}     & 34.1 & 27.5 & 25.5 & 51.0 & 29.0 & 39.0 & 48.0 & 40.5 & 38.0 & 22.5 & 22.5 & 43.0 & 34.0 & 22.5 & 32.5 & 45.5 & 32.5 & 40.0 & 30.0 & 21.0 & 37.0 \\
    VideoChat \citep{videochat}        & 35.5 & 33.5 & 26.5 & 56.0 & 33.5 & 40.5 & 53.0 & 40.5 & 30.0 & 25.5 & 27.0 & 48.5 & 35.0 & 20.5 & 42.5 & 46.0 & 26.5 & 41.0 & 23.5 & 23.5 & 36.0 \\
    TimeChat \citep{Ren2023TimeChat}          & 38.5 & 40.5 & 36.0 & 61.0 & 32.5 & 53.0 & 53.5 & 41.5 & 29.0 & 19.5 & 26.5 & 66.5 & 34.0 & 20.0 & 43.5 & 42.0 & 36.5 & 36.0 & 29.0 & 35.0 & 35.0 \\
    Video-LLaVA \citep{video-llava}    & 43.0 & 46.0 & 42.5 & 56.5 & 39.0 & 53.5 & 53.0 & 48.0 & 41.0 & 29.0 & 31.5 & 82.5 & 45.0 & 26.0 & 53.0 & 41.5 & 33.5 & 41.5 & 27.5 & 38.5 & 31.5 \\
    P-LLaVA-7B \citep{pllava}          & 46.6 & 58.0 & 49.0 & 55.5 & 41.0 & 61.0 & 56.0 & 61.0 & 36.0 & 23.5 & 26.0 & 82.0 & 39.5 & 42.0 & 52.0 & 45.0 & 42.0 & 53.5 & 30.5 & 48.0 & 31.0 \\
    ShareGPT4Video \citep{sharegpt4video} & 51.2 & 49.5 & 39.5 & 79.5 & 40.0 & 54.5 & 82.5 & 54.5 & 32.5 & \textbf{50.5} & 41.5 & 84.5 & 35.5 & 62.5 & 75.0 & 51.0 & 25.5 & 46.5 & 28.5 & 39.0 & 51.5 \\
    ST-LLM \citep{st-llm}              & 54.9 & 66.0 & 53.5 & 84.0 & 44.0 & 58.5 & 80.5 & 73.5 & 38.5 & 42.5 & 31.0 & 86.5 & 36.5 & 56.5 & 78.5 & 43.0 & 44.5 & 46.5 & 34.5 & 41.5 & 58.5 \\
    VideoGPT+ \citep{videogpt+}        & 58.7 & 69.0 & 60.0 & 83.0 & 48.5 & 66.5 & 85.5 & 75.5 & 36.0 & 44.0 & 34.0 & \textbf{89.5} & 39.5 & \textbf{71.0} & \textbf{90.5} & 45.0 & 53.0 & 50.0 & 29.5 & 44.0 & 60.0 \\
    VideoChat2~\cite{mvbench} & 60.4 & 75.5 & 58.0 & 83.5 & 50.5 & 60.5 & 87.5 & 74.5 & \textbf{45.0} & 47.5 & 44.0 & 82.5 & 37.0 & 64.5 & 87.5 & 51.0 & 66.5 & 47.0 & \textbf{35.0} & 37.0 & \textbf{72.5} \\

    \midrule
    VideoChat2-textualized-task & 64.8	& 76.5 & 56.0 & \textbf{88.5}	 	& \textbf{52.5} & 77.0 & \textbf{92.5} & 74.0 & 41.0 & 	\textbf{50.5}	& 		45.0 & 87.0 &47.0 & \textbf{74.0} & 89.0 & 48.0 &85.0 & 	45.0	& 34.0	& 58.5		& \textbf{73.0}				\\%

    \midrule 
    VideoChat-TPO &\textbf{66.8} & \textbf{84.0} & \textbf{69.5} & 87.5 & 52.0 & \textbf{77.0} & 92.0 & \textbf{81.0} & 40.5 & 42.5 & \textbf{55.0} & 89.0 & \textbf{47.5} & 68.0 & 89.0 & \textbf{58.0} & \textbf{87.0} & \textbf{57.5} & 27.0 & \textbf{60.0} & 72.0 \\
    \bottomrule 
\end{tabular}
}
\caption{\textbf{Results on MVBench Multi-choice Question Answering.}}
\label{atab:results_mvbench}
\end{table*}

\begin{table*}[t!]
\centering

\resizebox{0.99\textwidth}{!}
{
\begin{tabular}{l|lllllllllllllllllllllllllll}
\toprule
Model & Overall & CR & ER & FD & FC & SC & VCor & VQA & VGR & FR & HR & I2IR & MIC & PR & S2IR & STD & STS & T2IR & VR & AQA & GAR & MVU & MEV & NIP & TL & TO & VidCap \\
  &   & GuAR & GNAP & TC & VClz & VCo & VO & EVQA & HE & IQASC & ICSC & ISTE & ITRSC & MAR & MR & JPS & 3DE & 3DOD & 3DOT & 3DPE & 3DSR & 3DQA & PT & RPM & SOT & 3DCR & 3DIR \\
\midrule

OpenFlamingo~\cite{awadalla2023openflamingo} & 22.3 & 25.5 & 25.8 & 24.6 & 21.6 & 25.0 & 28.2 & 34.5 & 49.0 & 14.5 & 19.0 & 13.5 & 22.5 & 17.5 & 26.0 & 39.0 & 49.0 & 20.0 & 27.5 & 10.0 & 13.5 & 16.5 & 30.0 & 20.0 & 18.7 & 24.5 & 22.5 \\
  &   & 25.0 & 21.5 & 25.5 & 25.0 & 14.5 & 13.5 & 15.5 & 27.5 & 4.0 & 25.5 & 23.0 & 7.0 & 22.1 & 3.0 & 1.5 & 26.5 & 22.0 & \textbf{35.0} & 17.0 & 28.5 & 20.5 & 23.5 & 11.5 & 31.0 & 25.0 & 23.5 \\
\midrule
XComposer2~\cite{ixc2} & 21.9 & 24.0 & 21.0 & 10.8 & 5.8 & 0.0 & 0.0 & 34.2 & 24.0 & 14.5 & 2.5 & 23.0 & 63.5 & 19.0 & 26.0 & 14.5 & 31.0 & 9.5 & 28.5 & 31.5 & 59.5 & 44.0 & 30.0 & 4.5 & 15.5 & 12.0 & 66.0 \\
  &   & 55.0 & 35.0 & 42.5 & 22.5 & 2.5 & 19.0 & 20.0 & 8.0 & 15.5 & 45.0 & 0.0 & 0.0 & 20.6 & 0.0 & 16.5 & 0.0 & 7.0 & 0.0 & 4.5 & 0.0 & 33.5 & 63.0 & 1.5 & 38.5 & 42.0 & 33.0 \\
\midrule
Qwen-chat~\cite{bai2023qwen} & 15.9 & 20.5 & 2.5 & 13.3 & 2.5 & 9.9 & 5.9 & 31.2 & 23.8 & 10.5 & 19.5 & 12.5 & 41.0 & 5.5 & 13.5 & 29.5 & 45.0 & 3.0 & 12.0 & 10.0 & 52.5 & 18.5 & 16.5 & 2.5 & 3.6 & 5.5 & 47.0 \\
  &   & 29.0 & 23.0 & 18.0 & 6.0 & 6.0 & 6.0 & 32.0 & 9.0 & 13.5 & 17.0 & 15.5 & 3.5 & 40.2 & 15.8 & 16.5 & 16.5 & 22.5 & 17.5 & 13.0 & 14.5 & 14.0 & 8.0 & 3.0 & 8.5 & 1.5 & 0.5 \\
\midrule

LLaVA-v1.5~\cite{llava} & 19.2 & 14.1 & 4.2 & 13.7 & 5.8 & 1.9 & 6.9 & 27.3 & 35.0 & 6.5 & 12.5 & 12.5 & 53.0 & 10.0 & 25.5 & 66.5 & 43.0 & 19.0 & 3.5 & 2.5 & 23.5 & 36.5 & 12.0 & 16.5 & 6.7 & 7.0 & 28.0 \\
  &   & 24.5 & 17.5 & 40.0 & 15.0 & 21.5 & 4.0 & 26.0 & 7.5 & 26.5 & 17.5 & 5.0 & 4.5 & 25.6 & 27.1 & 8.5 & 8.0 & 4.0 & 6.0 & 6.0 & 14.5 & 29.5 & 66.0 & 2.0 & 35.0 & 34.5 & 28.5 \\
\midrule
ShareGPT4V~\cite{chen2023sharegpt4v} & 18.5 & 16.4 & 5.0 & 10.8 & 6.2 & 9.0 & 2.7 & 34.2 & 28.5 & 4.5 & 10.5 & 3.5 & 57.0 & 4.0 & 12.5 & 55.5 & 44.5 & 13.5 & 5.0 & 5.0 & 26.0 & 38.0 & 14.0 & 15.5 & 10.9 & 6.0 & 25.0 \\
  &   & 26.5 & 19.0 & 42.0 & 7.5 & 14.0 & 7.5 & 31.5 & 7.0 & 29.0 & 18.0 & 5.0 & 1.5 & 28.1 & 23.3 & 9.5 & 3.0 & 7.0 & 6.0 & 2.0 & 8.0 & 27.5 & 65.5 & 0.0 & 44.0 & 36.5 & 31.0 \\
\midrule

LLaVA-interleave~\cite{li2024llava} & 32.4 & 29.5 & 24.8 & 26.3 & 23.2 & 26.4 & 25.1 & 48.8 & 49.8 & \textbf{23.5} & 25.0 & 28.0 & 57.0 & \textbf{21.5} & \textbf{33.0} & 63.5 & \textbf{54.5} & 25.0 & 26.0 & 24.0 & 27.0 & 49.5 & 29.0 & 23.0 & 25.4 & \textbf{27.5} & 32.5 \\
  &   &  43.0  &  34.0  &  49.0  &  \textbf{29.5}  &  \textbf{32.0}  &  \textbf{26.0}  &  30.0  &  21.5  &  42.0  &  47.5  &  22.5  &  14.0  &  23.6  &  32.3  &  17.5  & \textbf{28.5} &  23.0  &  17.5  &  3.0  &  31.0  &  36.0  & \textbf{79.0} &  \textbf{15.0}  &  \textbf{60.5}  &  34.5  &  42.5 \\
\midrule
InternVL1.5-chat~\cite{chen2023internvl} & 37.4 & 63.7 & \textbf{31.0} & 22.6 & 20.3 & 16.3 & 28.3 & \textbf{63.2} & 38.5 & 21.0 & \textbf{28.0} & 26.5 & 82.5 & 20.5 & 31.5 & 6.0 & 45.5 & 26.5 & 29.5 & 29.5 & 85.0 & 65.0 & 32.0 & 23.5 & 29.0 & 18.5 & 89.0 \\
  &   & \textbf{90.5} & 35.5 & 56.5 & 23.5 & 31.0 & 24.5 & 53.0 & 26.0 & 40.0 & \textbf{49.0} & 25.5 & 15.5 & 59.3 & 43.6 & 19.5 & 22.5 & 23.5 & 15.0 & \textbf{33.5} & 28.0 & 39.0 & 71.0 & 9.5 & 46.5 & \textbf{50.5} & 39.5 \\
\midrule

VideoChat2~\cite{mvbench} & 35.0 & 46.8 & 27.5 & 31.6 & 23.6 & 25.6 & \textbf{28.8} & 45.3 & 54.3 & 20.5 & 25.5 & 25.5 & 64.0 & 21.0 & 31.0 & 31.5 & 50.0 & 21.0 & \textbf{31.0} & \textbf{30.5} & 73.0 & 51.0 & 31.5 & 23.5 & 21.8 & 24.0 & 81.5 \\
  &   & 54.0 & \textbf{42.0} & \textbf{59.0} & 23.0 & 30.5 & 23.0 & 44.5 & \textbf{26.5} & \textbf{44.0} & 36.5 & 25.0 & \textbf{18.0} & 38.6 & \textbf{44.4} & 21.0 & 26.5 & 24.0 & 13.0 & 0.0 & \textbf{28.5} & \textbf{43.0} & 65.5 & 11.5 & 58.0 & 36.0 & 35.0 \\
\midrule

VideoChat-TPO & \textbf{40.2} & \textbf{73.3} & 24.3 & \textbf{37.0} & \textbf{24.6} & \textbf{26.5} & 26.9 & 45.0 & \textbf{69.5} & 20.5 & 23.5 & \textbf{29.5} & \textbf{83.0} & 21.0 & 31.0 & \textbf{92.5} & 
 49.5 & \textbf{29.5} & 30.0  & 24.5 & \textbf{88.0} & \textbf{67.5} & \textbf{34.5} & \textbf{29.5} & \textbf{36.8} & 24.5 & \textbf{94.5} \\
  &   & 59.0 & 39.5 & 56.5 & 27.5 & 29.5 & 21.0 & \textbf{59.0} & 25.0 & \textbf{44.0} & 48.5 & \textbf{27.5} & 14.5 & \textbf{73.4} & \textbf{44.4} & \textbf{23.5} & 27.5 & \textbf{24.5} & 7.5 & 0.0 & 24.0 & 38.5 & 67.0 & 11.5 & 58.5 & 47.0 & \textbf{40.5} \\

  \bottomrule
\end{tabular}
}
\caption{\textbf{Quantitative results of MMIU~\cite{meng2024mmiu}.} Accuracy is the metric, and the Overall score is computed across all tasks.  
}
\label{tab:fullMMIU}
\end{table*}

\begin{table}[ht]
  \centering
    \resizebox{\linewidth}{!}{
    \begin{tabular}{l|ccc|c|c}
    \toprule
    \multirow{2}{*}{Model} &\multicolumn{4}{c}{Charades-STA~\cite{charades-sta}} &\multicolumn{1}{|c}{MVBench} \\
    \cmidrule(lr){2-5} \cmidrule(lr){6-6}
    &R@0.3 &R@0.5 &R@0.7 &mIoU & AVG\\
    \midrule
        \ModelName   & \textbf{58.3} & \textbf{40.2} & \textbf{18.4} & \textbf{38.1} & \textbf{66.8}\\  

     ~~~~~~Only \small{QVHighlight}  &  54.8 & 34.6 & 15.1 & 35.8 & 66.5\\  
     
    \bottomrule 
\end{tabular}
}
\vspace{-3mm}
\caption{\textbf{Ablation task datasets.}}
\label{tab:ablate_tvg_supp}
\end{table}

\paragraph{MVbench.}
We present the detailed performance of MVBench in Table~\ref{atab:results_mvbench}, VideoChat-TPO achieves an average score of 66.8, increasing by 6.4 points based on VideoChat2. It gets superior performance among MLLMs with the same number of input frames and LLMs of comparable model scale.
In Action Localization, temporal labels in the VideoChat2-Textualized-Task are defined as text. While the model demonstrates strong capabilities in zero-shot temporal grounding, converting the task into a QA problem does not improve performance. However, by optimizing with TPO, the model can benefit from original label supervision, resulting in corresponding performance enhancements. Also, Its superior performance is particularly evident in tasks that require moment-based perception and reasoning, including Action Sequence (AS), Action Localization (AL) and Action Prediction (AP), with scores of 84.0 (+7.5\%), 55.0 (+10\%), and 69.5 (+13.5\%) respectively. This demonstrates the excellent potential of TPO in sophisticated video understanding tasks. 

\paragraph{MMIU.}
The results are shown in Table~\ref{tab:fullMMIU}. VideoChat-TPO shows a significant improvement over VideoChat2, achieving an overall score of 40.2 (+5.2\%). Compared with VideoChat2, Our model has achieved clear improvements in Causality Reasoning (CR), Visually Grounded Reasoning (VGR), Multiple Image Captioning (MIC), Spot the Difference (STD), General Action Recognition (GAR), Temporal Localization (TL), Video Captioning (VidCap), Multiview Action Recognition (MAR), Image Captioning with Spatial Context (ICSC), and Egocentric Video Question Answering (EVQA), 
with scores of 73.0 (+26.5\%), 69.5 (+15.2\%), 83.0 (+19\%), 92.5 (+61\%), 88.0 (+15\%), 94.5 (+13.0\%), 73.4 (+35.8\%), 48.5 (+12\%) and 59.0 (14.5\%), respectively. Among them, we suppose the improvement of TL capability comes from the optimization of our temporal head, and the improvement of VGR, STD, MAR and ICSC capabilities comes from the optimization of our region head and mask head. The enhancements observed in captioning, specifically in metrics such as MIC, IC, and VidCap, indicate an improvement of TPO to capture detailed visuals. 
Meanwhile, we find that the improvement in multi-image capabilities stems from enhanced instruction comprehension. Compared with video assessments, which primarily consist of multiple-choice questions, multi-image evaluations emphasize the accuracy of responses to specific questions. After optimization with TPO, the model has significantly improved its instruction following, leading to a higher success rate.
\paragraph{How Scaling Task Data Affect MLLMs.}
We perform an ablation experiment on the dataset of stage 2 to evaluate the impact of the task training data on the model performance. Specifically, we reduce the number of temporal grounding datasets from six to one (QVHighlight~\cite{qvhighlight}). As shown in Table~\ref{tab:ablate_tvg_supp}, using only one dataset leads to slightly worse conversational performance (-0.3\%) on MVBench and significantly poorer expert task performance (-5.6\%) on Charades-STA R@0.5, when compared to employing multiple temporal grounding datasets for training the temporal task head. Notably, this approach remains more effective than training after textualizing the task data in QA tasks like MVBench. This finding indicates that scaling task data gives notable performance improvements in both multimodal and specific vision tasks.
Various datasets are necessary for effectively enhancing TPO's dialogue capabilities and achieving zero-shot generalization to fine-grained visual tasks.

\paragraph{LLaVA-OV-TPO Performance on Video Benchmarks.}
\begin{table}[ht]
  \vspace{-2mm}
  \centering
    \resizebox{\linewidth}{!}{
    \begin{tabular}{l|ccc}
    \toprule
    Model & MVBench & VideoMME & PerceptionTest \\ 
    \midrule
        LLaVa-OV  & 56.7 & 58.2 & 57.1  \\  
        LLaVa-OV-TPO  & \textbf{64.8}(+8.1) & \textbf{61.3}(+3.1) & \textbf{64.0}(+6.9) \\  
    \bottomrule 
\end{tabular}
}
\vspace{-3.5mm}
\caption{\textbf{Perfermance of LLaVA-OV on Video Benchmarks.}}
\label{tab:rebuttal_ov}
\end{table}
According to Table~\ref{tab:rebuttal_ov},  TPO method demonstrates performance improvements on LLaVA-OV~\cite{llavaonevision} across multiple video benchmarks as it does in VideoChat~\cite{mvbench} model. Since TPO uses extra visual cues to guide MLLM, LLaVA-OV-TPO achieves an average score of 64.8(+8.1\%) on MVBench~\cite{mvbench} and 64.0(+6.9\%) on PerceptionTest~\cite{perceptiontest}. The notable improvement clearly demonstrates the model's greatly enhanced ability to perceive visual details. Moreover, LLaVA-OV-TPO achieves a 3.1\% performance improvement on VideoMME~\cite{videomme} and shows that the model has also made progress in knowledge modeling and understanding long videos.
These results suggest TPO method is effective across various models and is particularly beneficial for fine-grained perception tasks.

\section{Training and Data Details}

\begin{table*}[ht]
\centering
\resizebox{0.8\linewidth}{!}{
\begin{tabular}{l|cccc}
\toprule
Config & Stage 1 & Stage 2 & Stage 3 w/o Con. & Stage 3 \\
\hline
Vision Enc.  LR & Frozen & Frozen & 2e-5 & 2e-5 \\
Connector  LR & Frozen & Frozen & 2e-5 & 2e-5 \\
Temporal Head  LR & - & 1e-4 & 2e-5 & 2e-5 \\
Region Head  LR & - & 1e-4 & 2e-5 & 2e-5 \\
Mask Head  LR & - & Frozen & Frozen & Frozen \\
Mask Adapter  LR & - & 1e-4 & 2e-5 & 2e-5 \\
Temporal Token  LR & - & 2e-4 & 2e-5 & 2e-5 \\
Region Token  LR & - & 1e-4 & 2e-5 & 2e-5 \\
Mask Token  LR & - & 1e-4 & 2e-5 & 2e-5 \\
LLM LoRA  LR & 2e-5 & 2e-5 & 2e-5 & 2e-5 \\
LR Schedule & Cosine Decay & Cosine Decay & Cosine Decay & Cosine Decay \\
Optimizer & AdamW~\cite{loshchilov2017adamw} & AdamW~\cite{loshchilov2017adamw} & AdamW~\cite{loshchilov2017adamw} & AdamW~\cite{loshchilov2017adamw} \\
Weight Decay & 0.02 & 0.02 & 0.02 & 0.02 \\
Input Resolution & 224$^2$ & 224$^2$ & 224$^2$ & 224$^2$ \\
Input Frames & 16 & 16 & 16 & 16 \\
LLM LoRA Rank & 16 & 16 & 16 & 16 \\
LLM LoRA Alpha & 32 & 32 & 32 & 32 \\
Warmup Ratio & 0.2 & 0.2 & 0.2 & 0.2 \\
Total Batch Size & 128 & 64/128/128 & 256 & 256 \\
Epoch & 1 & 25/3/1 & 1 & 3 \\
Numerical Precision & DeepSpeed bf16~\cite{rasley2020deepspeed} & DeepSpeed bf16~\cite{rasley2020deepspeed} & DeepSpeed bf16~\cite{rasley2020deepspeed} & DeepSpeed bf16~\cite{rasley2020deepspeed} \\
\bottomrule 
\end{tabular}
}
\caption{\textbf{Training Settings of VideoChat-TPO.} Con. means conversation data and LR means learning rate.} 
\label{atab:train_config}
\end{table*}
\begin{table*}[ht]
\centering
\resizebox{0.8\linewidth}{!}{
    \begin{tabular}{l|c|c|c}
    \toprule
    \textbf{Stage} & \textbf{Task} &\textbf{Samples} & \textbf{Datasets}\\
    \midrule 
    \multirow{3}{*}{Stage 1} 
    & Temporal Grounding & 50K &  DiDeMo~\cite{DiDeMo}, QuerYD~\cite{queryd} \\
    & Spatial Grounding & 50K & RefCOCO~\cite{yu2016refcoco}, RefCOCOg~\cite{yu2016refcoco}, RefCOCO+~\cite{mao2016refcocog} \\
    & Segmentation & 50K & SAMv2~\cite{ravi2024sam2}, MeViS~\cite{mevis} \\
    \midrule 
    \multirow{3}{*}{Stage 2} 
    & Temporal Grounding & 116.5K & DiDeMo~\cite{DiDeMo}, QuerYD~\cite{queryd}, HiRest~\cite{hirest}, ActivityNet~\cite{caba2015activitynet}, TACoS~\cite{regneri2013grounding}, NLQ~\cite{grauman2022ego4d}  \\
    & Spatial Grounding & 540.0K & AS-V2~\cite{wang2024allseeing_v2}, Visual Genome~\cite{krishna2017vg}, RefCOCO~\cite{yu2016refcoco}, RefCOCO+~\cite{yu2016refcoco}, RefCOCOg~\cite{mao2016refcocog} \\
    & Segmentation & 114.6K & SAMv2~\cite{ravi2024sam2}, MeViS~\cite{mevis} \\
    \midrule 
    \multirow{4}{*}{Stage 3} 
    & Temporal Grounding  & 7.5K & QVHighlight~\cite{qvhighlight} \\
    & Spatial Grounding & 400K & AS-V2~\cite{wang2024allseeing_v2}, Visual Genome~\cite{krishna2017vg}, RefCOCO~\cite{yu2016refcoco}, RefCOCO+~\cite{yu2016refcoco}, RefCOCOg~\cite{mao2016refcocog} \\
    & Segmentation & 116.5K & MeViS~\cite{mevis}, SAMv2~\cite{ravi2024sam2} \\
    & Temporal Reasoning & 40K & YouCook2~\cite{DaXuDoCVPR2013}, ActivityNet~\cite{caba2015activitynet} \\
    & Conversation &3M & VideoChat2-IT~\cite{mvbench}, ShareGPT-4o~\cite{chen2023internvl}, LLaVA-Hound-DPO~\cite{zhang2024direct},  ShareGPT4V~\cite{sharegpt4v}\\
\bottomrule 
\end{tabular}
}
\caption{\textbf{Datasets Used at Three Training Stages.} The temporal grounding task includes two subtasks: moment retrieval and highlight detection.}
\label{tab:data}
\end{table*}

Table~\ref{atab:train_config} and \ref{tab:data} lists the detailed training configurations and data of VideoChat-TPO in different stages. In each stage, the model is parametrized from the weights from the previous stage and continues training. 

\paragraph{Settings of Stage 1.} The LLM is equipped with LoRA~\cite{hu2021lora} for saving computational memory, using a LoRA rank of 16 and an alpha of 32. Only the LoRA is trained for efficiency. We adopt the AdamW optimizer~\cite{loshchilov2017adamw} with the peak learning rate of 2e-5 and use cosine weight decay. The training involves a total batch size of 128 across 32 A100 GPUs.
Since the purpose of stage 1 is to make MLLM identify tasks, we only use a small amount of data in this stage and adopt LLM loss so that LLM can generate task-specific tokens. For each task, we train the LLM with 50k examples to recognize the task.
For training data, we use DiDeMo~\cite{DiDeMo} and QuerYD~\cite{queryd} for temporal grounding task, RefCOCO~\cite{yu2016refcoco}, RefCOCOg~\cite{yu2016refcoco} and RefCOCO+~\cite{yu2016refcoco} for spatial grounding task, and SAMv2~\cite{ravi2024sam2}, MeViS~\cite{mevis} for segmentation task.

\paragraph{Settings of Stage 2.} In stage 2, we add the task heads (\ie temporal head, region head, and mask head) and learnable task tokens (temporal token, region token, and mask token). The objective of the second training stage is to learn the task head with preliminary functional capabilities. Therefore, we train LLM, task head and task token at this stage, and freeze vision encoder and connector.

In stage 2, the region head and token are trained with a learning rate of 2e-5 using a cosine learning rate scheduler. We use a two-layer MLP as region head to train from scratch and we use MSE loss for region head training. For training data, we use AS-V2~\cite{wang2024allseeing_v2}, Visual Genome~\cite{krishna2017vg}, RefCOCO~\cite{yu2016refcoco}, RefCOCOg~\cite{yu2016refcoco}, RefCOCO+~\cite{yu2016refcoco} for one epoch with a total batch size of 128 to train region head and token.

We use a learning rate of 1e-4 for the temporal head and 2e-4 for the temporal token in stage 2. The temporal head is the same as CG-DETR~\cite{moon2023correlation} in structure, but we use the pre-trained InternVideo2~\cite{wang2024internvideo2} to extract video features, while query features are extracted using the Chinese-Llama-Alpaca~\cite{chinese-llama-alpaca}. We use the same loss function in CG-DETR.
We train the model on DiDeMo~\cite{DiDeMo}, QuerYD~\cite{queryd}, HiRest~\cite{hirest}, ActivityNet~\cite{caba2015activitynet}, TACoS~\cite{regneri2013grounding}, NLQ~\cite{grauman2022ego4d} for 25 epochs with a total batch size of 64.

For the mask head, we use the pre-trained SAM2~\cite{ravi2024sam2} model, replacing the prompt encoder of SAM2 with a single MLP layer called the mask adapter.  During training, the mask token and adapter are trained with a learning rate of 2e-5, and the rest of SAM2 is frozen. We use MeViS~\cite{mevis}
, SAMv2~\cite{ravi2024sam2} for three epochs in this stage with a total batch size of 128. We supplement the training data by expanding the ASv2~\cite{wang2024allseeing_v2} image dataset into videos and adding it to this stage.

\paragraph{Settings of Stage 3.}
The third training stage aims to strengthen the model's conversational ability using TPO. This stage is divided into two parts. The first part involves training on a combined dataset of all tasks. The second part uses a dataset combining both task and conversation data. For conversatation data, we use VideoChat2-IT~\cite{mvbench}, ShareGPT-4o~\cite{chen2023internvl},  LLaVA-Hound-DPO~\cite{zhang2024direct},  ShareGPT4V~\cite{chen2023sharegpt4v} for instruction finetuning. We adopt a peak learning rate of 2e-5 for all the model in this stage and use a total batch size of 128.

\paragraph{TPO Additional Training  and Inference Cost.}

\begin{table}[ht]
  \vspace{-2mm}
  \centering
    \resizebox{\linewidth}{!}{
    \begin{tabular}{l|c|ccc}
    \toprule
    Model & GPU & Stage1 & Stage2 & Stage3 \\ 
    \midrule
        VideoChat-TPO   & 64 & 0.5h & 11h & 52h \\  
       ~~~~~~textualized task data  & 64 & 0.5h & 10h & 50h \\  
       ~~~~~~only conversation data  &  64 & - & - & 42h \\  
    \bottomrule 
\end{tabular}
}
\vspace{-3.5mm}
\caption{\textbf{Training Cost of Three Stages on VideoChat.} Textualized task data means converting task data into conversation form.}
\label{tab:training_cost}
\end{table}
From the data perspective, as can be seen from Table~\ref{tab:data} in the Appendix, we have very little training data in the first (around 0.15M) and second stages (around 0.7M), most of the data (around 3.5M) is used in third phase of the experiment. Among the data in the third stage, most of it is conversation data for fine-tuning MLLM. Therefore, TPO introduces little new data. Concerning training cost, according Table ~\ref{tab:training_cost}, when using the same amount of data, the training time of our TPO method and the autoregressive method is almost the same, and compared with the version without visual task, the TPO method increases the training cost by about 25\%.

The Temporal Head and the Mask Head contains additional encoders. In training phase, the additional encoders are frozen, and we use the features extracted by the encoder for training. In inference phase, the additional encoders are only used when the task head is activated. When only performing conversation tasks, no additional inference cost is incurred.

\begin{table*}[ht]
    \centering
    
    \begin{minipage}[t]{1.0\linewidth}
        \centering
        \begin{tcolorbox}[colback=white!100]
            \footnotesize
            \begin{tabular}{p{\linewidth} c}
                1. Localize the visual content described by the given textual query $\langle query \rangle$ in the video, and output the start and end timestamps in seconds. \\
                2. Detect and report the start and end timestamps of the video segment that semantically matches the given textual query $\langle query \rangle$. \\
                3. Locate and describe the visual content mentioned in the text query $\langle query \rangle$ within the video, including timestamps. \\
                4. The given natural language query $\langle query \rangle$ is semantically aligned with a video moment, please give the start time and end time of the video moment. \\
                5. Find the video segment that corresponds to the given textual query $\langle query \rangle$ and determine its start and end seconds. \\
            \end{tabular}
        \end{tcolorbox}
        \vspace{-3mm}
        \caption{\textbf{Instructions for Temporal Grounding.}}
        \label{tab:template_temoral}
    \end{minipage}

    \begin{minipage}[t]{1.0\linewidth}
        \centering
        \begin{tcolorbox}[colback=white!100]
            \footnotesize
            \begin{tabular}{p{\linewidth} c}
                1. Track the object in the video using a box with initial coordinates $\langle track\_box \rangle$.\\
                2. Use a bounding box with coordinates $\langle track\_box \rangle$ to follow the movement of the moving object in the visual input.\\
                3. Given an initial bounding box with coordinates $\langle track\_box \rangle$, track the motion of the target object in the sequence of frames.\\
                4. Starting from the box defined by the coordinates $\langle track\_box \rangle$, monitor the movement of the object in the video.\\
                5. Utilizing the initial box specified by the coordinates $\langle track\_box \rangle$, continuously track and update the location of the object in the video stream.\\
                6. Given a video with an object of interest enclosed in a bounding box with coordinates $\langle track\_box \rangle$, generate a sequence of bounding boxes that track the object's movement.\\
                7. With an initial box defined by $\langle track\_box \rangle$, trace the object's trajectory by generating a sequence of bounding boxes that follow the object's movement in the visual input.\\
                8. Apply an object tracking algorithm to a video, starting with a bounding box defined by $\langle track\_box \rangle$.\\
                9. Given a video and an initial bounding box defined by $\langle track\_box \rangle$, track the movement of the object within the video.\\
                10. Starting from an initial box defined by $\langle track\_box \rangle$, track the movement of the object in the visual input.\\
            \end{tabular}
        \end{tcolorbox}
        \vspace{-3mm}
        \caption{\textbf{Instructions for Tracking.}}
        \label{tab:template_tracking}
    \end{minipage}

    \begin{minipage}[t]{1.0\linewidth}
        \centering
        \begin{minipage}[t]{0.4\linewidth}
        \centering
        \begin{tcolorbox}[colback=white!100]
        \footnotesize
        \begin{tabular}{p{\linewidth} c}
        1. Where is $\langle expr \rangle$?\\
        2. Can you find $\langle expr \rangle$?\\
        3. Can you detect $\langle expr \rangle$?\\
        4. Can you locate $\langle expr \rangle$?\\
        5. Please find $\langle expr \rangle$.\\
        6. Please detect $\langle expr \rangle$?\\
        7. Please locate $\langle expr \rangle$?\\
        8. Find $\langle expr \rangle$.\\
        9. Detect $\langle expr \rangle$?\\
        10. Locate $\langle expr \rangle$?\\
        \end{tabular}
        \end{tcolorbox}
        \vspace{-3mm}
        \caption{\textbf{Instructions for Spatial Grounding.}}
        \vspace{-3mm}
        \label{tab:template_region}
        \end{minipage}%
        \hfill
        \begin{minipage}[t]{0.5\linewidth}
        \centering
        \begin{tcolorbox}[colback=white!100]
        \footnotesize
        \begin{tabular}{p{\linewidth} c}
        1. Please give the motion path of $\langle obj \rangle$ in the video over time.\\
        2. Show the tracking trajectory of $\langle obj \rangle$'s movement through the scene in the video.\\
        3. Please generate a motion path of $\langle obj \rangle$'s movement in the video, highlighting its tracking trajectory.\\
        4. Show the tracking trajectory of $\langle obj \rangle$.\\
        5. Generate $\langle obj \rangle$'s tracking trajectory.\\
        6. Visualize the tracking trajectory of $\langle obj \rangle$ in the video.\\
        7. Please generate a visual representation of $\langle obj \rangle$'s movement in the video, highlighting its tracking trajectory.\\
        \end{tabular}
        \end{tcolorbox}
        \vspace{-3mm}
        \caption{\textbf{Instructions for Referring Segmentation.}}
        \vspace{-3mm}
        \label{tab:template_segmentation}
        \end{minipage}
    \end{minipage}

\end{table*}

\paragraph{Template Details.}

To support the proper invocation of task-specific decoders, we construct a series of instruction templates for different tasks and use them as instruction tuning data for MLLM. We comprehensively list all the instruction templates below, in Table~\ref{tab:template_temoral}, \ref{tab:template_tracking}, \ref{tab:template_region}, and \ref{tab:template_segmentation}.

\section{Qualitative Results}

\begin{figure*}[t]
    \centering
    \includegraphics[width=\linewidth]{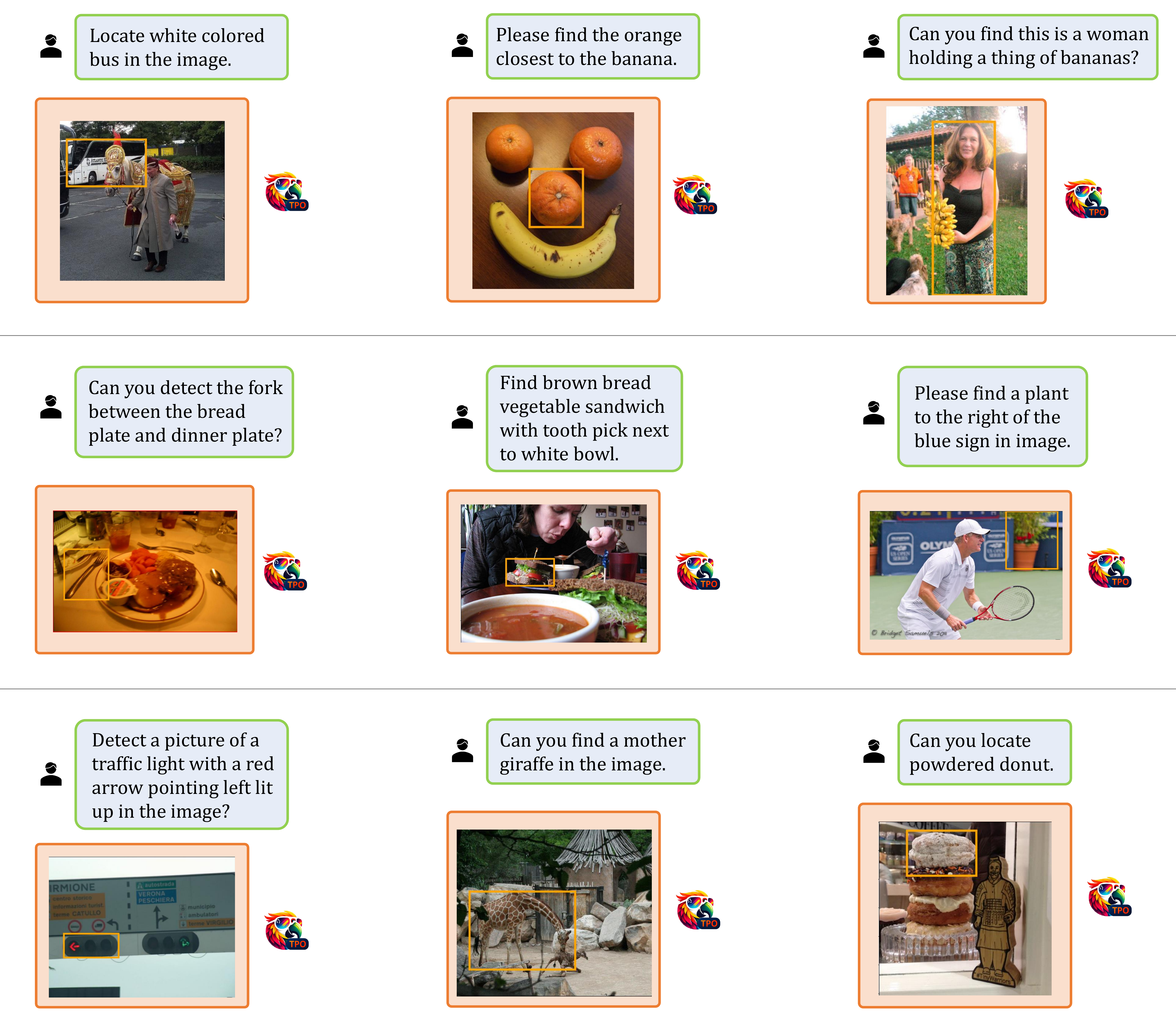}
    \caption{\textbf{Qualitative Results of Spatial Grounding.}}
    \label{fig:sg}
\end{figure*}

\begin{figure*}[t]
    \centering
    \includegraphics[width=0.95\linewidth]{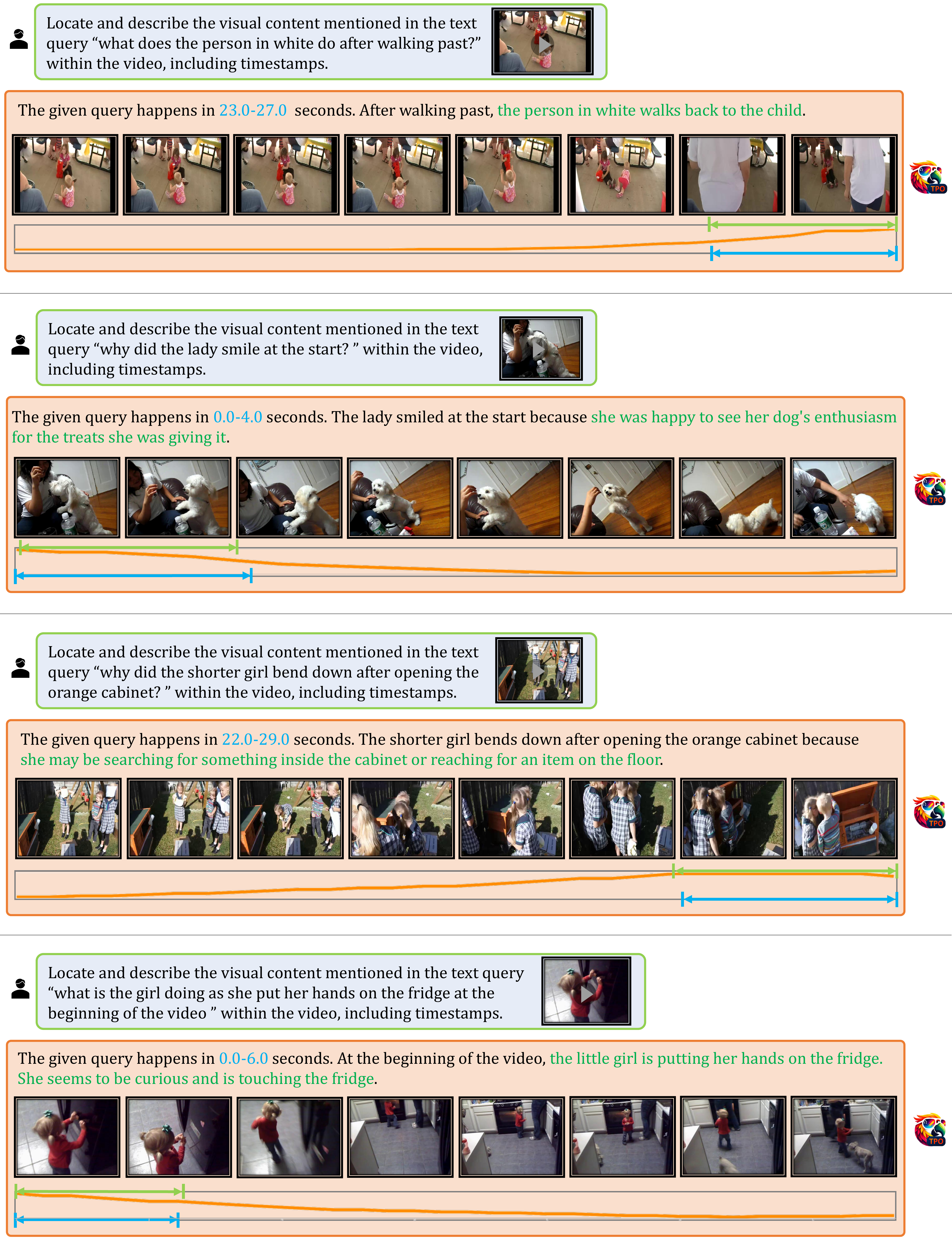}
    \caption{\textbf{Qualitative Results of Moment Retrieval and Highlight Detection.} The orange curve represents saliency score, the blue interval represents the time interval predicted by the model, and the green interval represents the ground truth.}
    \label{fig:mr&hl}
\end{figure*}

\begin{figure*}[t]
    \centering
    \includegraphics[width=\linewidth]{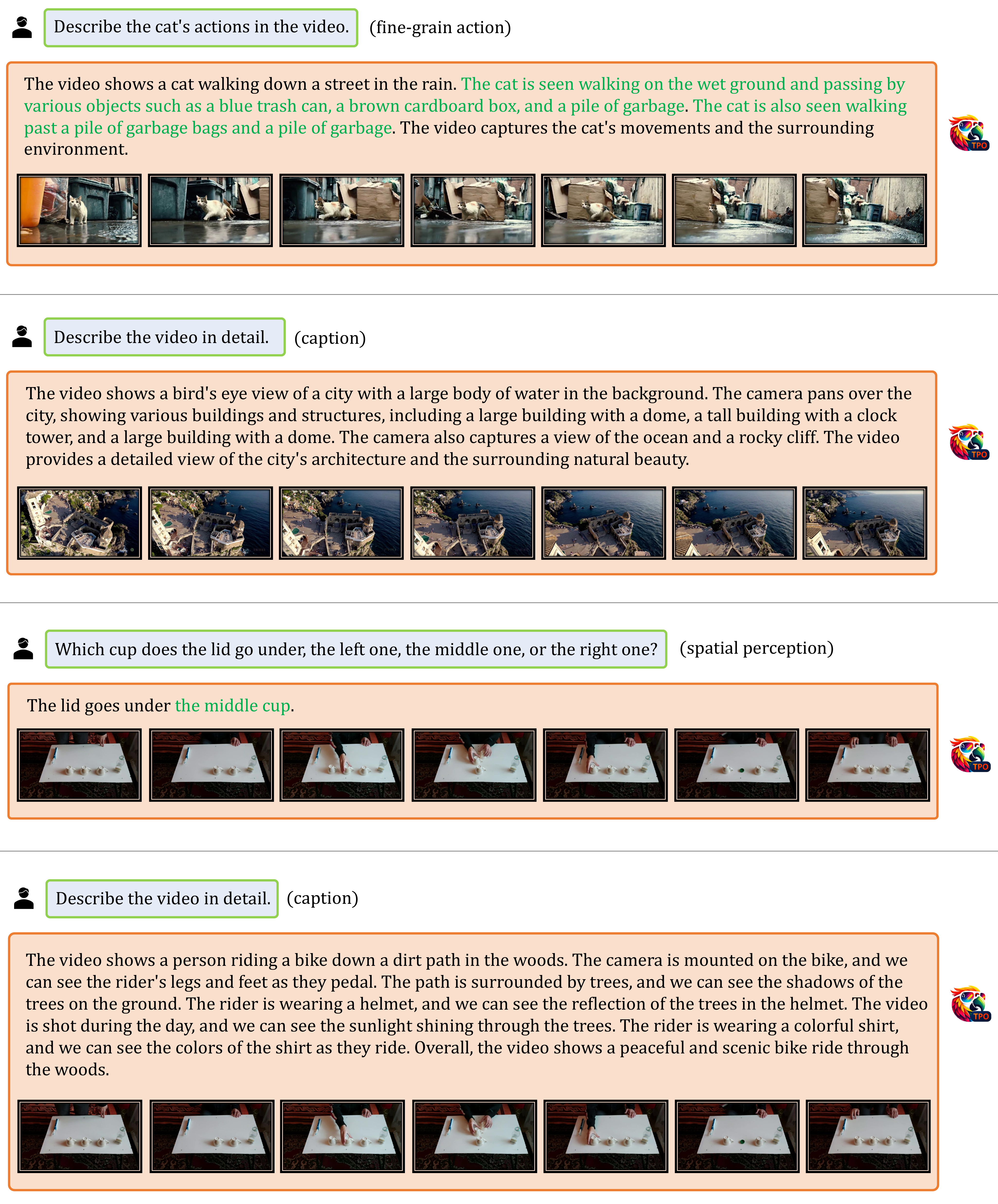}
    \caption{\textbf{Qualitative Results of Multimodal Video Understanding.}}
    \label{fig:conv}
\end{figure*}

We evaluate VideoChat-TPO on various visual perception tasks and display the visualizations from Figure~\ref{fig:sg} to Figure Figure~\ref{fig:mr&hl}. In addition, we also show the results of multimodal video understanding in Figure~\ref{fig:conv}.

\vspace{-5mm}
\paragraph{Spatial Grounding.}
In Figure~\ref{fig:sg}, we show the spatial grounding visualizations. VideoChat-TPO can infer the target object from the description of natural language and locate it. Our VideoChat-TPO can accurately locate the target among multiple similar objects. Even if the target object is occluded or in the background area, it can still be accurately located.

\begin{figure*}[t]
    \centering
    \includegraphics[width=\linewidth]{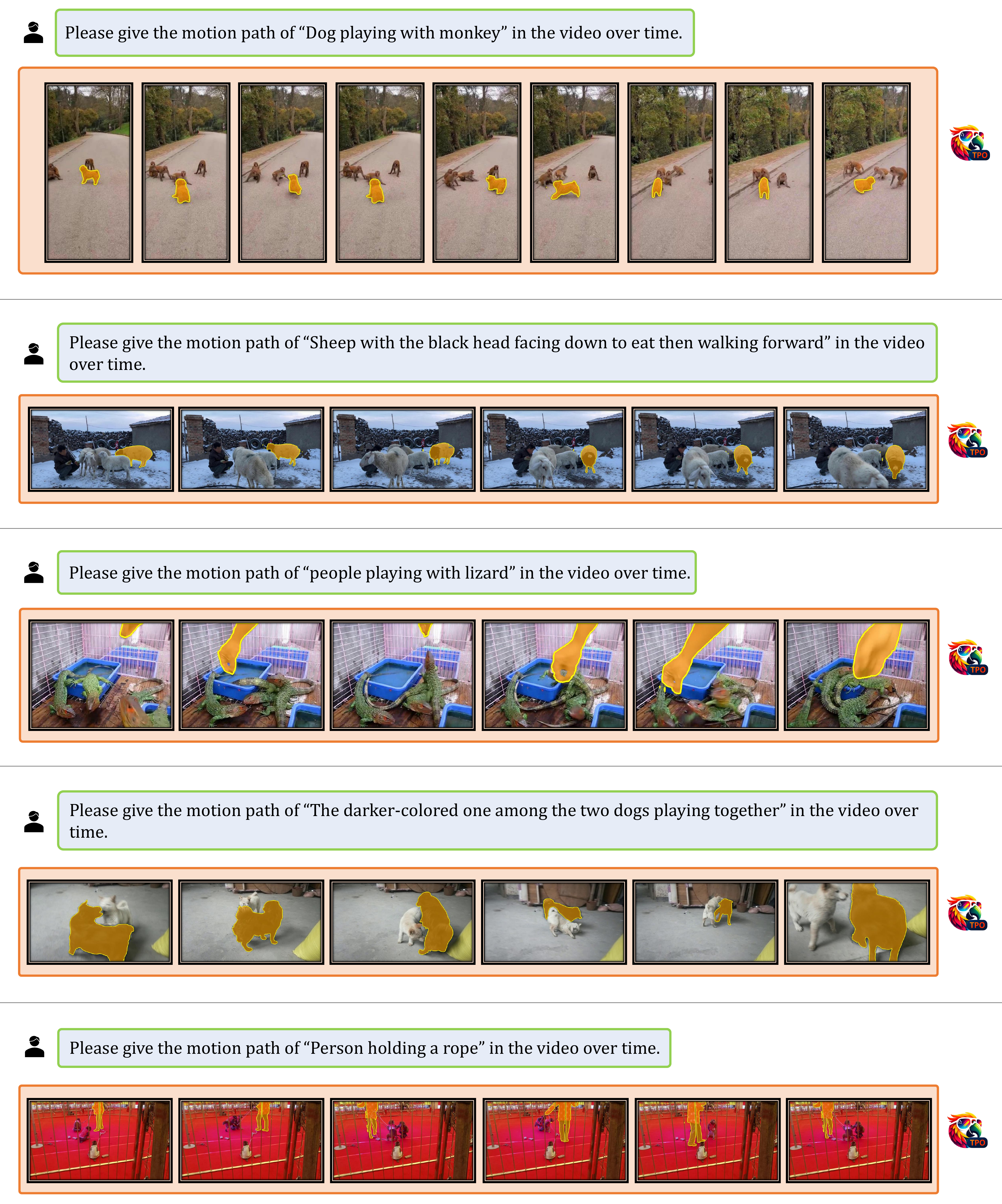}
    \caption{\textbf{Qualitative Results of Referring Segmentation.}}
    \label{fig:seg}
\end{figure*}

\vspace{-5mm}
\paragraph{Referring Segmentation.}
We show the visualizations of the referring segmentation in Figure~\ref{fig:seg}. VideoChat-TPO can delinear the target object in the video according to user input in complex scenes. Furthermore, VideoChat-TPO can separate the target object from multiple objects of the same kind according to the description of appearance or action characteristics indicated by the user.

\begin{figure*}[t]
    \centering
    \includegraphics[width=\linewidth]{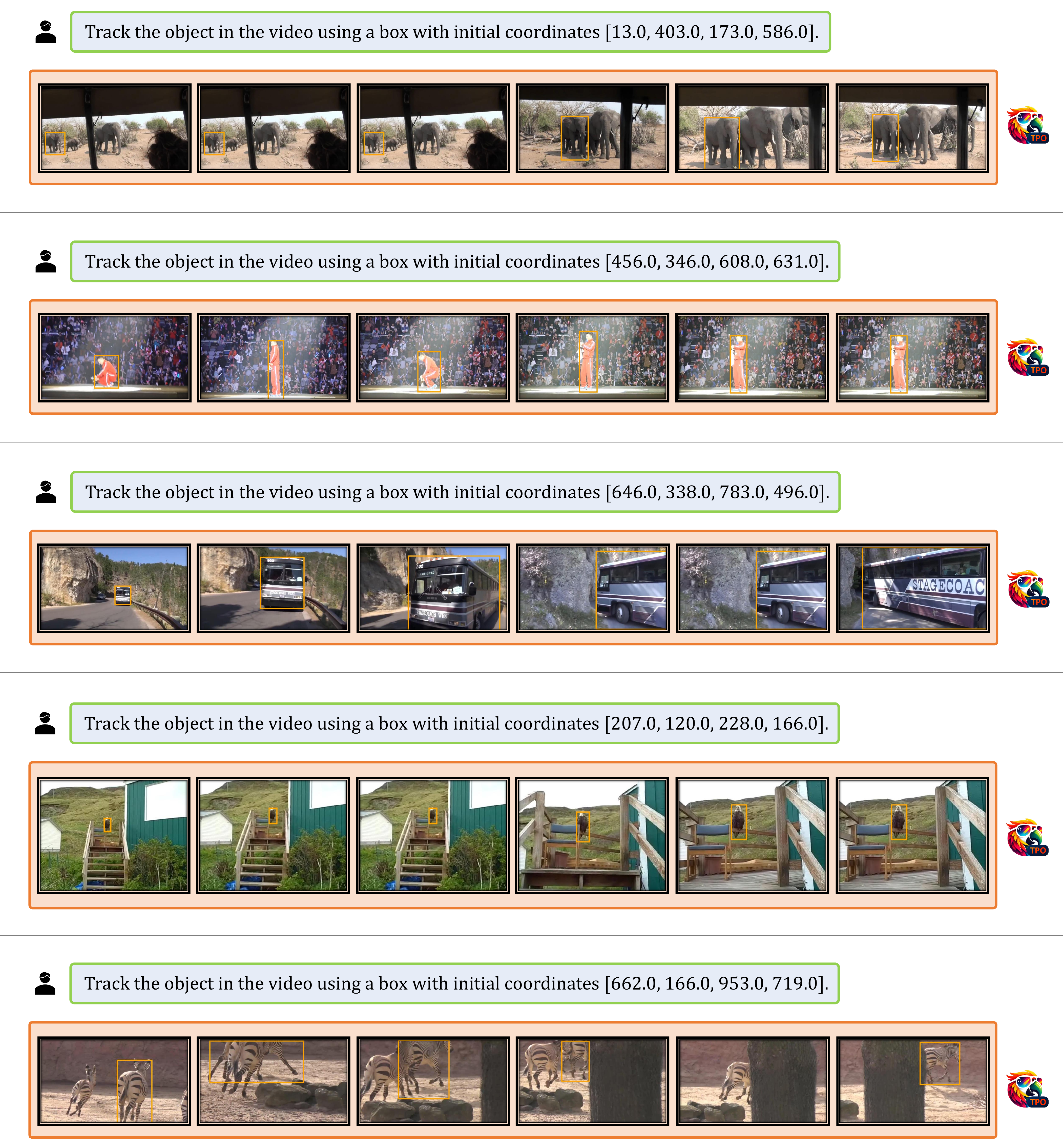}
    \caption{\textbf{Qualitative Results of Tracking.}}
    \label{fig:tracking}
\end{figure*}

\vspace{-5mm}
\paragraph{Tracking.}
The tracking visualizations are shown in Figure~\ref{fig:tracking}. The user needs to include the bounding box coordinate information of the first frame of the tracked target in the video in the input. The visualizations show that when the target object is partially occluded in the video, it can still be tracked. Even if the target object is out of the camera's view, our VideoChat-TPO can still track it when it appears in subsequent frames.

\paragraph{Moment Retrieval and Highlight Detection.}
The visualizations of the moment retrieval and highlight detection are given in Figure~\ref{fig:mr&hl}. Our VideoChat-TPO can infer the results and target events based on the user's questions, and perform moment retrieval and highlight detection on the target events.

\vspace{-5mm}
\paragraph{Multimodal Video Understanding.}
The multimodal video understanding visualizations are shown in Figure~\ref{fig:conv}. Our VideoChat-TPO achieve decent results in fine-grained action description, spatial description, and video captioning.

\end{document}